\documentclass[preprint,5p,twocolumn,sort&compress]{elsarticle}
\usepackage{amsmath,mathtools}
\usepackage{multirow}
\usepackage[english]{babel} % English language/hyphenation
\usepackage{bm} % Math packages
\usepackage{amsthm}

\usepackage{subfig}
\usepackage{hyperref}

\usepackage[Symbol]{upgreek}
\usepackage{bm}
\newcommand{\vct}{\mathbf}
\usepackage{comment}
\usepackage{graphics} % for pdf, bitmapped graphics files
\usepackage{epsfig} % for postscript graphics files
\usepackage{mathptmx} % assumes new font selection scheme installed
\usepackage{times} % assumes new font selection scheme installed
\usepackage{amsmath} % assumes amsmath package installed
\usepackage{amssymb}  % assumes amsmath package installed
\usepackage{enumerate}
\usepackage[x11names]{xcolor}
\usepackage{float}
\usepackage{fancyhdr}

\usepackage{nonfloat}
\PassOptionsToPackage{normalem}{ulem}
\usepackage{ulem}
\providecolor{added}{rgb}{0,0,0}
\providecolor{deleted}{rgb}{1,0,0}
\newcommand{\added}[1]{{\color{added}{}#1}}

\newtheorem{theorem}{Theorem}

\newtheorem{remark}{Remark}
\usepackage{capt-of}

\usepackage{tikz}
\usetikzlibrary{calc}

\fancypagestyle{firstpage}
{
    \fancyhead[L]{© 2022. This manuscript version is made available under the CC-BY-NC-ND 4.0 \href{https://creativecommons.org/licenses/by-nc-nd/4.0/}{ https://creativecommons.org/licenses/by-nc-nd/4.0/}
        \textbf{\\}
    \textbf{\\}
    This manuscript has been accepted for publication in Robotics and Autonomous Systems.  The formal publication can be accessed through  DOI: \href{https://doi.org/10.1016/j.robot.2022.104073}{10.1016/j.robot.2022.104073}}   
    \cfoot{}
    \fancyfoot[L]{\small{\textit{Accepted  by Robotics and Autonomous Systems, Elsevier}}}
 
}

\begin{document}

\let\WriteBookmarks\relax
\def\floatpagepagefraction{1}
\def\textpagefraction{.001}
\begin{frontmatter}

\title{
 A passive admittance  controller to enforce Remote Center of Motion and Tool Spatial constraints  with application in hands-on surgical procedures
}
\thispagestyle{firstpage}

\author{Theodora Kastritsi $^*$}\ead{tkastrit@ece.auth.gr}

\author{Zoe Doulgeri}\ead{doulgeri@ece.auth.gr}

\address{Department of Electrical and Computer Engineering, Aristotle University of Thessaloniki, Thessaloniki
54124, Greece}

\tnotetext[t1]{This  research  is  co-financed  by  Greece  and  the  European  Union  (European  Social  Fund-  ESF)  through  the  Operational  Programme  "Human Resources  Development,  Education  and  Lifelong  Learning"  in  the  context of  the  project  "Strengthening  Human  Resources  Research  Potential  via Doctorate Research"(MIS-5000432), implemented by the State Scholarships Foundation (IKY).}

%%%%%%%%%%%%%%%%%%%%%%%%%%%%%%%%%%%%%%%%%%%%%%%%%%%%%%%%%%%%%%%%%%%%%%%%%%%%%%%%
\begin{abstract}
The restriction of feasible motions of a manipulator link constrained to move through an entry port is a common problem in minimum invasive surgery procedures. Additional spatial restrictions are required to ensure the safety of sensitive regions from unintentional damage.  
In this work, we design a target admittance model that is proved to enforce robot tool manipulation by a human through a remote center of motion and to guarantee that the tool will never enter or touch forbidden regions.
The control scheme is proved passive under the exertion of a human force ensuring manipulation stability, and smooth natural motion in hands-on surgical procedures enhancing the user’s feeling of control over the task. Its performance is demonstrated by  experiments with a setup mimicking a hands-on surgical procedure comprising a KUKA LWR4+ and a virtual intraoperative environment. 

\end{abstract}

\begin{keyword}
  Physical Human-Robot Interaction \sep RCM manipulation \sep active constraints \sep surgical robots \sep variable damping
\end{keyword}

\end{frontmatter}
\thispagestyle{firstpage}

%%%%%%%%%%%%%%%%%%%%%%%%%%%%%%%%%%%%%%%%%%%%%%%%%%%%%%%%%%%%%%%%%%%%%%%%%%%%%%%%
\section{Introduction}
The vision of having robots work
collaboratively with humans is slowly beginning to materialize in both industrial and professional robotics. In physical
human-robot interaction (pHRI), humans can bring experience, knowledge, perception
and understanding for the proper execution of a task and
robots can reduce fatigue and increase human capabilities
in terms of strength, speed and accuracy.
There are tasks  where the last manipulator link is constrained to move through an entry port, thus being only allowed to translate and rotate along its axis and rotate about the entry point. The existence of such a remote center of motion (RCM) constraint, if it is not incorporated in the robot's controller, may lead to high cognitive load for the human as she/he has to account for them during the interaction. For example, in hands-on  minimally  invasive surgical procedures, the surgeon manipulates a long thin tool attached at the robot's end-effector passing through an incision point on the patient's body by exerting forces on the tool basis.  Any such task is even more demanding when there are spatial constraints regarding the tool tip or the whole tool. In the surgical case, these spatial constraints concern sensitive tissues like arteries and veins that should not be accidentally injured during the operation.  Hence the robot's controller should further guarantee the avoidance of these regions and preferably provide a haptic feedback when the human manipulates the tool close to them.   

In general, RCM and spatial constraints may either be real or desired.  For example, 
 in some surgical robots, the RCM constraint is achieved by mechanical means. 
 If the robot is however a general purpose manipulator, the RCM should be imposed by the control action to ensure minimum stress of the incision wall. Spatial task constraints may either be hard constraints like the existence of a wall or rigid surface or soft constraints  that should not be stressed by external forces according to the task. The latter is typical in surgical procedures in which forbidden region avoidance should be incorporated in the robot's controller as it is essential for the patient's safety.  Even in tasks with hard spatial constraints, actively respecting them would decrease the physical and cognitive human effort.    
 
 In this work, we propose an admittance controller
 satisfying both RCM  and forbidden region constraints during the robot's kinesthetic guidance. As we shall see in the next subsection, the existing works in the literature, have 
 addressed these two objectives separately either for autonomous or kinesthetic  guidance cases. Control schemes have been proposed for industrial and minimum invasive surgical tasks for spatial constraint satisfaction. In surgery spatial constraints are related to forbidden regions. These constraints  can be known analytically \cite{khatib1980commande, papageorgiou2020passive, theodorakopoulos2015impedance,bettini2002vision,bettini2004vision,bowyer2015dissipative}, can be generated utilizing Dynamic Movement Primitives that encode the  demonstration trajectory \cite{papageorgiou2020kinesthetic,papageorgiou2020passiveRCIM}  or provided as point clouds produced by the perception system \cite{ kastritsi2019guaranteed,ryden2012forbidden,ryden2013advanced,leibrandt2014implicit}.
 RCM enforcement schemes are mainly proposed for robotic assisted minimal invasive surgery (RAMIS) % [Dora ECC TRMB, de Momi etc] 
 \cite{kastritsi2020human,kastritsi21TRMB,SANDOVAL201895,sandoval2018generalized,su2019improved,Azimian10,yang2019adaptive,sadeghian2019constrained,9265418,Aghakhani13,boctor2004virtual,pham2015analysis,kapoor2006constrained,leibrandt2014implicit,he2014multi}. \added{The proposed target admittance model is designed in a way that decouples the robot’s joint space into the RCM constrained and unconstrained subspace so that RCM constraints are satisfied during tool manipulation and the control action that guarantees manipulation away from the forbidden regions  acts only
along the RCM unconstrained directions}. It is moreover proved passive under the exertion of a generalized human force which ensures manipulation stability in all cases. \added{The proposed target admittance model, incorporates a novel variable damping term   in order to achieve a smooth manipulation performance near the forbidden regions}. Our experimental results in  a virtual intraoperative environment, demonstrate that with our controller a user can effectively achieve intuitive RCM manipulation of a long tool away from forbidden areas providing haptic feedback in the form of repulsive forces when the tool is near them. 

\subsection{Related Works}\label{relatedWorks}
 
Active  constraints or virtual fixtures    were firstly  introduced in  tele-robotic  manipulation  by Rosenberg providing force feedback from virtual environments to reduce the cognitive load of the user  \cite{rosenberg1992use,rosenberg1993virtual}  and have been utilized for both hands-on and teleportation applications in surgical \cite{kastritsi2019guaranteed, bowyer2015dissipative,bettini2002vision,bettini2004vision,moccia2019vision,leibrandt2014implicit,petersen2013dynamic} industrial \cite{colgate2003intelligent, lin2006portability,papageorgiou2020kinesthetic,papageorgiou2020passive,papageorgiou2020passiveRCIM}  or even in underwater robotic tasks \cite{ryden2013advanced}. They can be classified as either virtual fixtures for enforcing barriers around forbidden regions \cite{kastritsi2019guaranteed,khatib1980commande,leibrandt2014implicit,ryden2012forbidden,ryden2013advanced,petersen2013dynamic} or  virtual fixtures for assisting guidance achieving an attractive behavior towards a desired path \cite{bowyer2015dissipative,bettini2002vision,bettini2004vision,moccia2019vision, papageorgiou2020kinesthetic,papageorgiou2020passive,papageorgiou2020passiveRCIM}. Notice that an attractive virtual fixture in a region can be assumed as a forbidden-region virtual fixture for its complementary space and vice versa.  Their enforcement is implemented using energy storage methods, such as artificial potential fields \cite{khatib1980commande, theodorakopoulos2015impedance,papageorgiou2020passive,papageorgiou2020passiveRCIM,kastritsi2019guaranteed,papageorgiou2020kinesthetic,moccia2019vision,leibrandt2014implicit,ryden2012forbidden,ryden2013advanced,petersen2013dynamic} or via controllers that do not store energy 
{\cite{bettini2002vision, bettini2004vision,bowyer2015dissipative}}. The latter approach does not guarantee constraint satisfaction in all cases  and is not able to provide haptic cues when the robot is not moving.  Artificial potentials are unbounded \cite{khatib1980commande,kastritsi2019guaranteed,theodorakopoulos2015impedance,papageorgiou2020passive,moccia2019vision} or bounded energy functions \cite{papageorgiou2020passiveRCIM,papageorgiou2020kinesthetic,ryden2012forbidden,ryden2013advanced,leibrandt2014implicit,petersen2013dynamic} depending on the specific objective they address. For example, penetrated (bounded) artificial potentials are utilized around trajectories encoded by Dynamic Movement Primitives to allow a user to inspect kinesthetically as well as significantly modify a previously learned path  \cite{papageorgiou2020passiveRCIM,papageorgiou2020kinesthetic}.  A method enforcing constraints utilizing unbounded functions are presented in  \cite{kastritsi2019guaranteed} and \cite{kastritsi2019manipulation} to guarantee that the robot tool tip and the whole tool respectively will never touch a forbidden surface provided by a point cloud. 
A  review  about artificial potential fields can be found in \cite{bowyer2013active}.

 Tool manipulation via an RCM  is typical in RAMIS where the tool is either  directly manipulated by the surgeon or indirectly via a telemanipulation set-up. The former known as hands-on robotic surgery is preferred in some cases \cite{petersen2013dynamic}. In hands-on robotic surgery the robot is under an impedance or admittance control scheme and the human force is directly applied on the tool. In teleoperated setups the surgeon manipulates a haptic device which generates velocities or displacements that are sent as reference velocities or positions respectively to the patient-side robot controller.

 Different control  approaches about the satisfaction of  the RCM constraint and tool tip  trajectory tracking  have been  proposed  for autonomous operation or teleoperation set-ups  ~\cite{Azimian10,SANDOVAL201895,Aghakhani13,boctor2004virtual,pham2015analysis, su2019improved, sandoval2018generalized, yang2019adaptive,sadeghian2019constrained,9265418}. 
Some of them require a torque level interface which is available in a limited number of the commercially available robotic manipulators  \cite{sandoval2018generalized, SANDOVAL201895, su2019improved}.     In  \cite{SANDOVAL201895} the proposed solution constrains 3-dof  instead of 2-dof that are required, by specifying a tool orientation compatible with the RCM given a desired tool-tip position trajectory. Unintentional external forces that may occur on the arm are mapped in the null space of the task. An optimization method is used in  \cite{boctor2004virtual} to find the optimal joint configuration that satisfies the RCM constraint as well as reaching of a tool-tip target point but the controller may be trapped in a local minimum. In \cite{pham2015analysis} a linear map is used, that transforms the velocity of the haptic device so that it satisfies the RCM constraint of the surgical robot. Works \cite{Azimian10,Aghakhani13} adopt a task priority  approach with the  RCM  constraint  being in  the  first priority  level and the tool-tip  trajectory  tracking in the  second. First order inverse kinematics are used and the approaches are validated only through simulations.  In \cite{sadeghian2019constrained}, \cite{yang2019adaptive} an inverse first order kinematic controller is designed to track a desired tip trajectory respecting the RCM constraint.  All the above works do not provide proofs of stability and  RCM constraint satisfaction.

There  are  limited  works  constraining  the  motion  of  the tool to satisfy a RCM in hands-on RAMIS procedures \cite{he2014multi, kapoor2006constrained, leibrandt2014implicit}. They also lack proof of closed loop system passivity and guarantees of  RCM constraint satisfaction. In \cite{kastritsi2020human}, we propose a control law applied at the torque level that required  accurate knowledge of the robot's dynamics and validate it through simulations. Its transfer to a real robot requires a robot with a torque interface and \added{accurate knowledge of the dynamic parameters}  which is difficult in the majority of cases. \added{To deal with this problem we have proposed an admittance control scheme in \cite{kastritsi21TRMB} ensuring passivity  and  manipulation  of  long  instruments  via  a  RCM}. Methods proposed for teleoperated or autonomous operation cannot be directly applied in hands-on procedures particularly those proposing first order inverse kinematic solutions. If the human force is used in place of desired velocities or displacements in these methods, the  zero target inertia which is implied negatively affects stability. In particular, as discussed in \cite{lamy2009achieving}, the target inertia is lower bounded in order for the system to remain passive in a realistic admittance control case.

To the best of our knowledge spatial constraints together with RCM tool manipulation is only addressed in \cite{9265418} and  \cite{leibrandt2014implicit}. The authors in \cite{9265418} propose a method to preserve the safety of  sensitive human organs  and achieve  tip tracking by monitoring the minimum distance from the sensitive area or setting a threshold on the force exerted on the sensitive area in order to stop the robot.  
In \cite{leibrandt2014implicit} the authors propose   a torque level control method for a RAMIS hands-on procedure. The RCM is achieved by regulating the tool to  a  desired orientation  that is calculated to be compatible with the RCM constraint thus constraining 3-dof instead of 2-dof. Active constraints  are imposed by producing repulsive forces on the tool shaft as the tool comes closer to the forbidden area defined by a point cloud. As the two controllers are superimposed,  system dynamics are coupled with each controller affecting the performance of the other. Moreover, no proof or guarantees are provided that the tool will not touch the forbidden area or it will not enter the empty space between  the points of the cloud.   
\added{ \subsection{Contributions}}
 
 In this work, we address the problem of satisfying
both RCM and forbidden region constraints via a novel admittance control design. A combination of our previous works \cite{kastritsi21TRMB} on RCM constraint   and forbidden region constraint satisfaction \cite{kastritsi2019guaranteed}, \cite{kastritsi2019manipulation} produces a system that is not passive and consequently it cannot guarantee stability. In fact, we faced various stability problems when we had initially experimented with this control combination. 
 We have thus redesigned the whole control scheme so that we can  successfully address both objectives of RCM tool manipulation and spatial constraint enforcement close to forbidden areas via haptic feedback.  The novel design is differentiated with respect to our previous works in several parts enumerated below. 
 
 \begin{enumerate}[(i)]
     \item The main novelty resides in the proposed mapping between the joint velocities and the free motion coordinates which differs from that of \cite{kastritsi21TRMB}. This novel mapping enables passivity  to  be  proved  in  the  presence  of  repulsive  potential and it allows free motion coordinates to be defined in  correspondence to the force components in a decoupled, independent way.
     The resulted motion is thus intuitive, enhancing transparency of manipulation. In contrast, the respective mapping in \cite{kastritsi21TRMB}, generates free coordinatecouplings leading to counter intuitive motion.
     \item In this work, the  null space basis of the RCM constraint Jacobian is  designed analytically as opposed to the algorithmic on-line solution utilized in  \cite{kastritsi21TRMB} which adds computational load to the overall solution and may result in discontinuities that affect manipulation performance.
     \item A novel variable damping term is incorporated in the target admittance dynamics proposed in this work, which is instrumental in achieving a smooth performance in the presence of the repulsive potentials that are introduced to enforce forbidden areas. As the introduction of such repulsive potentials  is equivalent to a non-linear  increase of the apparent stiffness in the vicinity of the constraints, the constant damping utilized in \cite{kastritsi21TRMB} jeopardizes performance.
     \item The method to enforce forbidden area constraints for the tool presented in our previous works \cite{kastritsi2019guaranteed}, \cite{kastritsi2019manipulation} does not consider RCM constraints and is implemented in the torque level via an impedance controller. In this work, we propose an admittance model which is shaped by the presence of the RCM constraints, so that repulsive forces are filtered. It is consequently not obvious that the sensitive areas will be protected. An essential contribution of this work is that we prove that active constraint satisfaction is guaranteed by the proposed scheme in all practical cases. 
 \end{enumerate}

Summarising, the contribution of this work is a novel admittance control scheme that:  
  \begin{itemize}
 \item guarantees stability of the overall system via passivity, achieving both objectives of RCM and sensitive area constraint satisfaction as testified by the theoretical proofs and validated by the experimental results
 
 \item provides  manipulation transparency and smooth motion in  hands-on surgical procedures  which enhances the user's feeling of control over the task.
 \end{itemize}

The rest of the paper is organized as follows. Section \ref{section:ProblemDescription} introduces  the problem and the control design objectives.  \added{Section \ref{background} summarises our background work on RCM constraint formulation and repulsive artificial potential fields for forbidden areas given as  point clouds.  Section \ref{section:ControlDesign} presents the proposed admittance control scheme  and proofs of passivity and control objectives achievement. The analysis  for the active constraint enforcement is extended to the whole tool body in subsection  \ref{extention} and the designed variable  damping is presented in  subsection \ref{damping}}. Experimental  results that demonstrate its effectiveness in a set-up mimicking a hands-on surgical procedure are presented  in  Section  \ref{section:experiment}. Conclusions are drawn in Section \ref{section:conclution}.

\section{Problem Description}\label{section:ProblemDescription}

Consider a general purpose robot with $n\ge 6$ degrees of freedom, which can be kinematically controlled that is common in most of the commercially available robotic equipment. The latter means that the robot is provided by an interface that accepts position or velocity reference commands and its internal controller ensures negligible errors in tracking them. Let a force/torque sensor be attached at its end-effector holding a long thin tool that is directly manipulated by a human through  an entry port.   Further consider the availability of a point cloud of an object  characterized as a forbidden  area that should not be touched by the tool. The task is the kinesthetic guidance of the tool to any accessible area through the port never touching the forbidden area. These type of tasks are typically encountered in hands-on robotic surgery.

The aim of this work is to design an admittance controller that would simultaneously achieve the following:
\begin{itemize}
  \item the entry port will be imposed as a RCM to the tool's manipulation by the guiding human 
  \item the tool-tip 
  %\deleted{$ \mathbf{p_t}(t) \in \mathbb{R}^3$} 
  or the tool body will never enter or even touch the forbidden area
    \item the user will feel repulsive forces in the vicinity of the forbidden area 
      
    \item  the controlled system will be passive with respect to the forces exerted by the user
    \item \added{a smooth tool manipulation performance is ensured close to forbidden areas }
 
\end{itemize}
The entry port may be either rigidly or softly attached to the surrounding environment as in the surgical applications. Then respectively, the first objective means that  constraint forces or entry port displacement will be  effectively zero during the task. The point cloud of the forbidden area is assumed to be provided by a camera registered in the robot's workspace like an endoscopic camera capturing intraoperative images in surgery. \added{Smooth tool manipulation refers to lack of tool oscillatory behaviour.}

\added{\section{Background} \label{background}
For completeness and clarity, the RCM constraint formulation and repulsive artificial potential are presented in this section, with more details to be found in our previous works \cite{kastritsi21TRMB} and \cite{kastritsi2019guaranteed}  respectively.}
\subsection{RCM constraint formulation}
\begin{figure}[!h]
  \centering
  \includegraphics[width=0.5\linewidth]{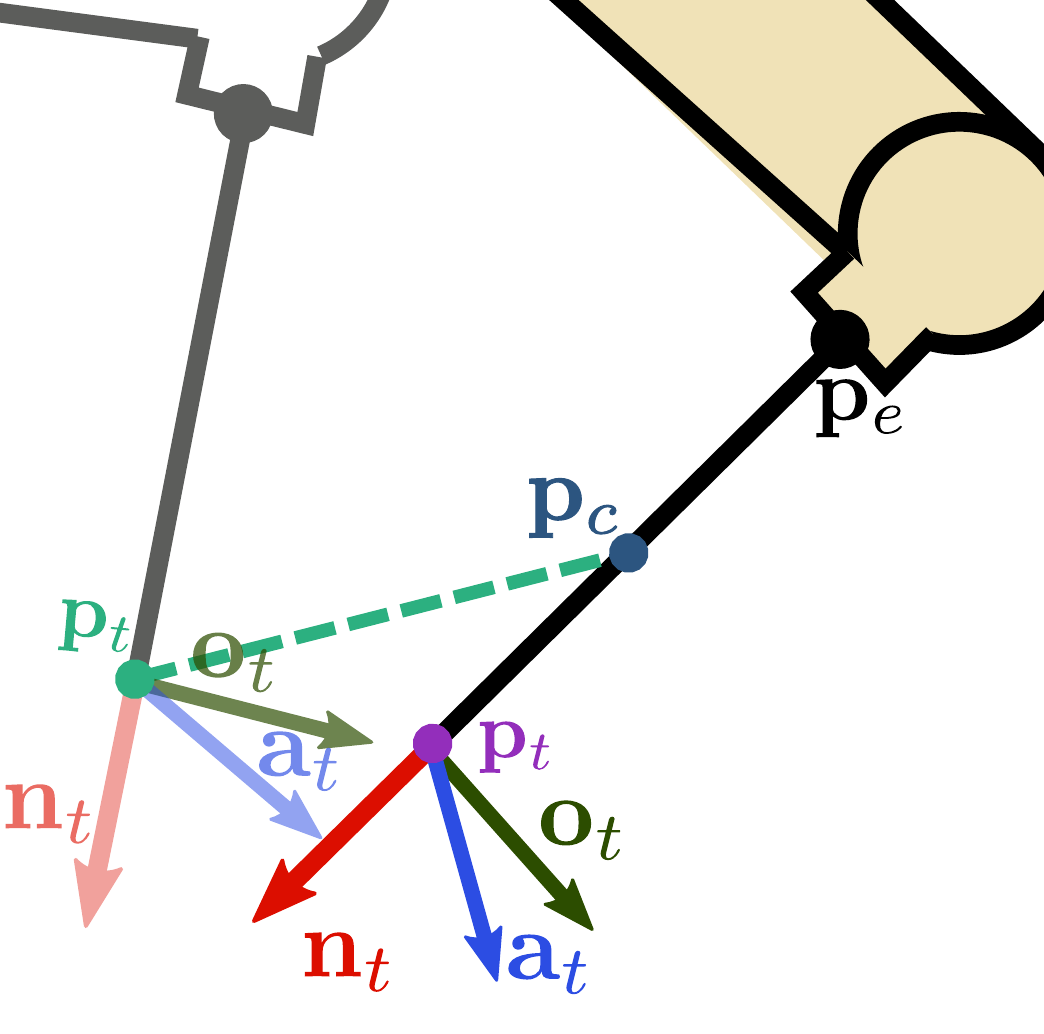}
  \caption{Two poses of the robot  \added{tool} with tip frames and the RCM point $\vct p_c$, passing through RCM (bold) and with an offset (grey).}
  \label{fig:kuka_tr}
\end{figure}

Let the position and the orientation of the tool-tip be given by $\vct p_t \in \mathbb{R}^{ 3} ,\ \vct R_t=[\vct a_t \ \vct o_t \ \vct n_t]  \in \mathbb{R}^{3\times 3} $ respectively.   Without loss of generality the unit  direction  of  the  tool  is assumed to be  $\vct n_t$.
 Let the  entry port position $\vct p_c\in \mathbb{R}^3$ be known. A way to find it, is for example, to manually guide the tool tip to the entry port to record its position in the robot's space.
 Given an initial robot configuration such that the tool axis $ \vct n_t $ passes through $\vct p_c$ we aim to impose 
$\vct p_c\ $as a RCM constraint during the manipulation of the tool by the user. Hence, the projection of $\vct p_t-\vct p_c$ on the plane orthogonal to $\vct n_t$ is desired to be kept close to zero at all times (see Figure \ref{fig:kuka_tr}). 
 Notice that $\vct B_c=[\vct a_t  \ \vct  o_t] \in \mathbb{R}^{3\times 2}$  is a basis of this plane since $\vct n_t^\mathrm{T} \vct B_c=\vct 0_{1\times2}$. Then, the desired RCM constraint is expressed as follows:
\begin{equation} \label{eq:postio_time_con}
   \vct x_c \triangleq \vct B^\mathrm{T}_c  (\vct p_t-\vct p_c)=\vct 0.
 \end{equation}
%  Notice that compare to our previous work [submitted paper] now we express the RCM constraint  with respect to the tool-tip frame instead of end effector frame.
 
 Taking the time derivative of \eqref{eq:postio_time_con}, yields the following  velocity constraint:
\begin{equation} \label{eq:vel_con}
   \dot{\vct x}_c= \vct B^\mathrm{T}_c \dot{\vct p}_t+\dot{ \vct B}^\mathrm{T}_c  (\vct p_t-\vct p_c)  =\vct 0. 
\end{equation}
By definition $\vct R_t=[\vct B_c  \ \vct n_t]$ and using the property $\vct{\dot R}_t=\bm{\widehat{\omega}}_t \vct{R}_t$, with $ \widehat{(.)}$ denoting the skew symmetric matrix of a 3-d vector, we get: $ \dot{ \vct B}_c=\bm{\widehat{\omega}}_t { \vct B}_c. $
Thus  \eqref{eq:vel_con} becomes:
 \begin{equation}\label{eq:pfidot}
\dot{\vct x}_c:=\vct A_x \vct v_t =\vct 0
\end{equation}
where $\vct A_x$ is the constraint Jacobian in the task space:
 \begin{equation}
\vct A_x=\vct B^\mathrm{T}_c [ \vct I_{3\times 3} \ \ {(\vct p_t-\vct p_c)^\wedge}]  \in \mathbb{R}^{2 \times 6}.
\end{equation}
Notice that by definition this is a full rank matrix (rank 2). 
Let $\vct{J_t}(\vct{q}_d) \in \mathbb{R}^{6 \times  n}$ be the robot Jacobian which maps the joint velocities $\vct{\dot q}_d$ to the generalized velocity of  the tool-tip  $\vct v_t=[\dot{\vct p}_t^\mathrm{T} \ \bm \omega_t^\mathrm{T}]^\mathrm{T} \in \mathbb{R}^6 $ i.e.:
\begin{equation}\label{eq:Jq}
   \vct v_t = \vct{J}_t(\vct{q}_d) \vct{\dot q}_d
\end{equation}
where $\dot{\vct p}_t, \ \bm\omega_t \in \mathbb{R}^3$ are the linear  and angular velocities of the tool-tip. We can therefore utilize \eqref{eq:Jq} to express \eqref{eq:pfidot} in the robot's joint space:
 \begin{equation}\label{eq:xc}
\dot{\vct x}_c:=\vct A\dot{\vct {q}}_d =\vct 0
\end{equation}
where
 \begin{equation}\label{eq:A}
\vct A=\vct A_x \vct J_t(\vct q_d)  \in \mathbb{R}^{2 \times n}
\end{equation}
denotes the constraint Jacobian in the joint space. Notice that matrix $\vct A$ is full rank assuming a motion away from kinematic singularities. 

\subsection{ Artificial potentials for active constraint enforcement}

Let the surface of the forbidden region be approximated   by a finite set of points $\mathcal{O}_s$ with positions $\vct p_i$. Let $\rho\in\mathbb{R}^+$ be the density of $\mathcal{O}_s$ in points per $cm^3$ which is considered  known and homogeneous. Then the side of a cube which includes one point is $\dfrac{1}{\sqrt[3]{\rho}}$. 
To cover the empty space   between the points in $\mathcal{O}_s$ we create spheres with radius $d_c=\frac{\sqrt{3}}{2\sqrt[3]{\rho}}$ 
centered at each point $\vct p_i$ so that each one contains one cube. We can thus guarantee empty space coverage by  overlapping spheres. We  then define  the  constrained surface as the boundary $\partial \mathcal{O}_c $ of the overlapping spheres  $\mathcal{O}_c $ (see Figure \ref{fig:sphe}): 
\begin{figure}[!h]
  \centering
  \includegraphics[width=0.64\linewidth]{ 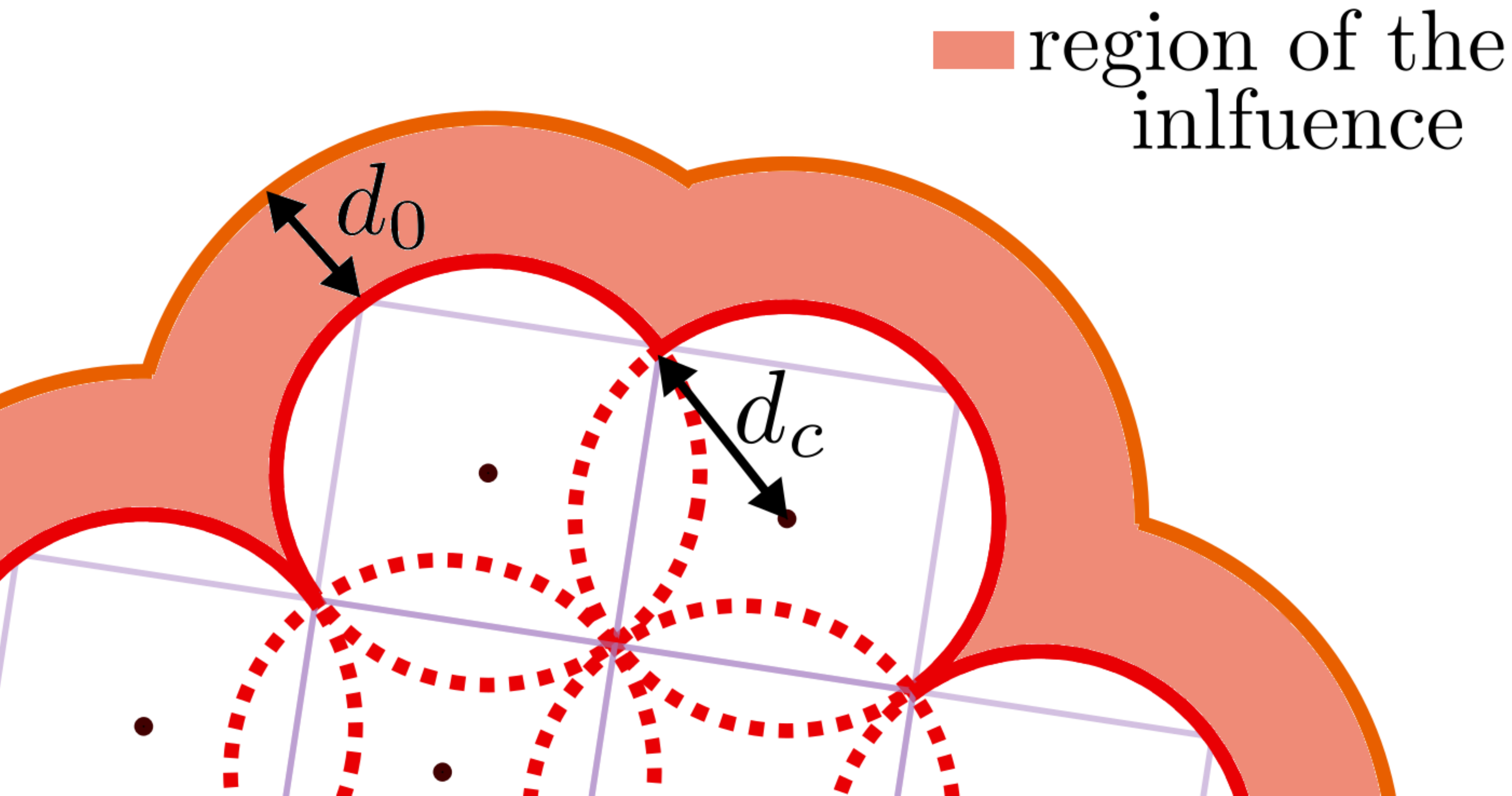}  \caption{Visualization of a cross section of the forbidden region with the overlapping spheres and the region of influence of the repulsive potential field. }
  \label{fig:sphe}
\end{figure}
\begin{equation}\label{eq:O_c}
\mathcal{O}_c = \bigcup_{\vct p_i \in \mathcal{O}_s}\{\vct{x}\in\mathbb{R}^3:   \|\vct{x} - \vct{p}_i\| \leq d_c \}.
\end{equation}

To ensure that the the boundary $\partial \mathcal{O}_c $ will never be penetrated or even touched we aim to impose a repulsive artificial potential with a predefined range of influence $d_0$ so that the user is repelled away from the forbidden area when the tip is approaching it by a distance less than $d_0$.

\begin{figure}[!h]
  \centering
  \includegraphics[width=0.9\linewidth]{ 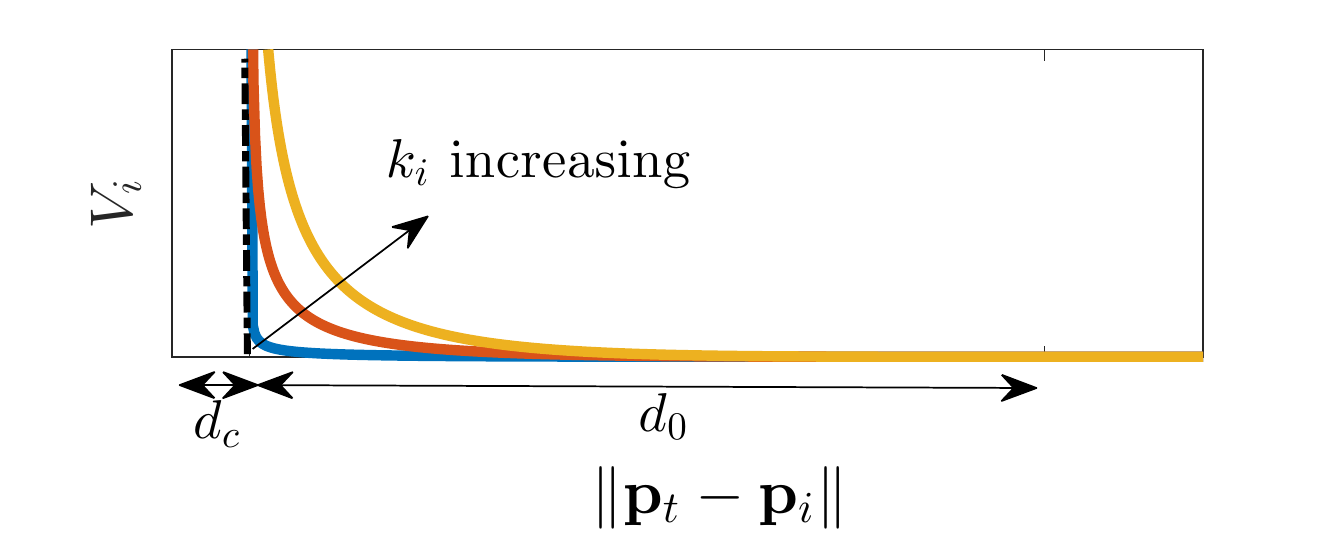}
  \caption{The repulsive artificial potential field \eqref{vi} with respect to the distance of the tool-tip  from a point of the cloud of the forbidden region.} 
  \label{fig:Vi}
\end{figure}

In particular, for each point of the cloud we utilize a field function $V_i(\vct p_t)$ with the following properties:
\begin{itemize}
\item $V_i(\mathbf{p}_t)=V(\|\vct p_t-\vct p_i\|)$ is a positive continuously differentiable scalar function, for all $ \|\mathbf p_t- \mathbf p_i\| \in (d_c, d_c+d_0] ;$
\item $V_i(\mathbf{p_t}) \rightarrow \infty$ if only if $\|\mathbf{p}_t - \mathbf{p}_i\|\rightarrow d_c ;$
\item  $\frac{\partial V_i(\mathbf{p}_t) }{\partial \mathbf{p}_t} $ is zero if and only if  $\|\mathbf{p}_t - \mathbf{p}_i\|\geq d_c+d_0$. 
\end{itemize}
In particular, the following function  introduced in our previous work  \cite{kastritsi2019guaranteed} is utilized: 
\begin{equation}\label{vi}
  V_i(\mathbf{p}_t)  =
  \begin{cases} 
       {\dfrac{k_i}{2} \ln\left(\frac{1}{1 - \psi_i}\right)^2,} &\quad \text{if} \  \|\mathbf{p}_t - \mathbf{p}_i\|\leq d_c+d_0, \\
       \text{0,} &\quad\text{else} \\
     \end{cases}
\end{equation}
where $k_i > 0$ is a scalar gain and
\begin{equation}\label{psi_i}
\psi_i = \frac{(\|\vct{p_t} - \vct{p}_i\| - (d_0+d_c))^2}{d_0^2}
\end{equation}
 This field function is depicted in Figure \ref{fig:Vi} for increasing values of $k_i$.  Notice that this repulsive potential field is zero beyond the range of influence and tends to infinity at the sphere boundary. Hence $V_i<\infty$  if and only if $\psi_i<1$ which is satisfied if and only if $\|\vct{p_t} - \vct{p}_i\|>d_c$.

The negative gradient  $ -\dfrac{\partial V_i}{\partial \vct{p_t}} $ of each $V_i$  \eqref{vi} produce repulsive forces at the tool tip $\vct{f}_i (\vct p_t)\in \mathbb{R}^3$:
\begin{equation}
  \vct{f}_i(\vct p_t) = 
  \begin{cases}
       {k_{vi}\vct{e}_i} &\quad \text{if} \  \|\mathbf{p}_t - \mathbf{p}_i\|\leq d_c+d_0, \\
       \text{0,} &\quad\text{else} \\
     \end{cases}\label{eq:f}
\end{equation}
where 
\begin{equation}
  \vct{e}_i =\left((d_0+d_c) - \|\vct{p}_t - \vct{p}_i\| \right)  \frac{\vct{p}_t - \vct{p}_i}{\|\vct{p}_t - \vct{p}_i\|}\in \mathbb{R}^3
\end{equation} is a vector with magnitude equal to the distance between the tool-tip and the sphere of radius $d_0+d_c$ centered at $\vct p_i$ with direction pointing from the center of the sphere towards the tool-tip, and 
% $k_{vi}\in \mathbb{R}$ given by:
\begin{equation}
    k_{vi}=\dfrac{2k_i}{d_0^2(1 - \psi_i)}\ln\left(\frac{1}{1 - \psi_i}\right) \in \mathbb{R}.
\end{equation} 
This scalar gain can be interpreted as a variable stiffness within the range of the field's influence which increases as the distance of tool-tip to the sphere surface decreases. Let the sum of repulsive forces produced by each artificial potential field $\vct{f}_i$ be:
\begin{equation} \label{eq:fr}
    \vct{f}_r =  \sum_{\vct p_i \in \mathcal{O}_s} \vct{f}_i.
\end{equation}
In  \cite{kastritsi2019guaranteed}, \eqref{eq:fr} is utilized to achieve enforcement of the constrained surface $\partial  \mathcal{O}_c$ via an impedance control scheme without RCM considerations.
\section{The Proposed \added{Admittance Control Scheme} }\label{section:ControlDesign}
\added{ Our aim is to design an admittance model that decouples the robot's joint space into the two spaces of the RCM constrained and unconstrained motion. To this aim,  we initially find a basis for the null space of the RCM constraint Jacobian, and then we proceed in the design of the desired target admittance model that is proved to fulfill all of the objectives set in section \ref{section:ProblemDescription}. Details on the extension of the forbidden area repulsive forces for the whole tool and the variable damping gains are given in separate subsections. 
}

\subsection{Null RCM constraint space}

We propose a basis for the null space of  $\vct A_x$ in which any forces should lie in order to avoid affecting the RCM constraint space. Let, this basis be denoted by matrix  $\vct Z_x\in \mathbb{R}^{4 \times 6} $ which should be composed  of vectors that are linearly independent and span the  null space of matrix $\vct A_x$ i.e. $\vct A_x \vct Z_x ^\mathrm{T}=\vct 0_{2\times 4}$. The following basis $ \vct Z_x $ for the null space of $\vct A_x$ is proposed:
   \begin{equation}\label{eq:Zx}
     \vct Z_x =   \begin{bmatrix}
    \ \ \ \ {\vct n}_t^\mathrm{T}  \ \ \ \ \ \vct 0_{1\times 3} \\
    (\vct p_t-\vct p_c)^\wedge \ \ \vct I_{3\times 3}
    \end{bmatrix}
   \end{equation}

Equivalently for the joint space we need to find a basis of the null space of the constraint Jacobian $\vct A$. Let this basis be denoted by 
$\vct Z \in \mathbb{R}^{(n-2)\times n}$ which should satisfy the following equation:
\begin{equation}\label{eq:AZ}
    \vct A \vct Z^\mathrm{T}=\vct 0\in \mathbb{R}^{2 \times (n-2)}.
\end{equation} 
 The following is a valid solution incorporating  $\vct Z_x$:  
% To express this basis with respect to the robot's joint space, we construct the null space of the constraint jacobian  $\vct A$ as:

    \begin{equation}\label{eq:Z}
     \vct Z=\begin{bmatrix}
     \vct Z_x (\vct J_t \vct J_t^\mathrm{T})^{-1}\vct J_t \\
     \vct G
     \end{bmatrix} \in \mathbb{R}^{(n-2) \times n} 
   \end{equation}
where  $\vct G \in \mathbb{R}^{(n-6) \times n} $ is the base matrix of the null space of $\vct J_t$ in case $n>6$ i.e. $\vct G \vct J_t^\mathrm{T}= \vct 0$  defined as in \cite{ott2008cartesian}.  We can easily verify \eqref{eq:AZ} for $\vct Z$ given by \eqref{eq:Z}.  

\added{\begin{remark} Notice that  the null space basis for the task in the joint space, $\vct Z_x 
(\vct J_t \vct J_t^\mathrm{T})^{-1}\vct J_t$,
is here designed analytically as opposed to the algorithmic on-line solution utilized in our previous work \cite{kastritsi21TRMB} which adds computational  load  to  the  overall  solution  but  more  importantly, it may result in discontinuities that affect manipulation performance.
\end{remark}}

A  general solution for the inverse  of the constraint motion kinematics $\dot{\vct x}_c=\vct A\dot{\vct {q}}_d $ 
% \eqref{eq:xc} 
is given by:
\begin{equation}\label{eq:qd:dot}
    \dot{\vct q}_d= \vct A^{\dagger}\dot{\vct x}_c+(\vct I_{n\times n}- \vct A^{\dagger}\vct A)\bm {\vct\xi}
\end{equation}
 where $ \vct A^{\dagger}=\vct W^{-1}\vct A^\mathrm{T}(\vct A \vct W^{-1} \vct A^\mathrm{T})^{-1}  \in \mathbb{R}^{n\times 2} $ is the weighted right pseudoinverse of $\vct A$ for a symmetric positive definite   matrix of weights $\vct W$ 
and $\bm {\vct\xi}\in  \mathbb{R}^{n}$ denotes an arbitrary vector. 
Denote the velocity of  the  unconstrained kinematics by $\dot{ \vct{x}}_f\in \mathbb{R}^{(n-2)}$. This velocity should be mapped in the joint space by $\vct Z^\mathrm{T}$ in order not to affect the constraint motion since $(\vct I_{n\times n}- \vct A^{\dagger}\vct A)\vct Z^\mathrm{T}=\vct Z^\mathrm{T}$. Thus, \eqref{eq:qd:dot} can be written in a compact form as:
\begin{equation}\label{eq:qd2}
    \dot{\vct q}_d= \vct S  \begin{bmatrix}
    \dot{\vct x}_c \\
    \dot{\vct x}_f
    \end{bmatrix}
\end{equation}
where $ \vct S  =[\vct A^\dagger  \  \vct Z^\mathrm{T} ] \in \mathbb{R}^{n\times n}. $ 
Let the weighted right pseudoinverse of   $\vct Z$ be defined by  $ \vct Z^{\dagger}=\vct W \vct Z^\mathrm{T}(\vct Z \vct W\vct Z^\mathrm{T})^{-1}  \in \mathbb{R}^{n\times (n-2)}$. Then the following properties hold between $\vct A$, $\vct Z$ and their pseudoinverses:
\begin{enumerate}[(i)]\label{listref}
    \item $\vct A \vct A^\dagger=\vct I_{2\times2}$
    \item $\vct Z^{{\dagger}^\mathrm{T}} \vct A^\dagger=\vct 0_{(n-2)\times 2}$
    \item $\vct Z \vct Z^{{\dagger}}= \vct Z^{{\dagger}^\mathrm{T}} \vct Z^\mathrm{T}=\vct I_{(n-2)\times(n-2)}$
\end{enumerate}

To recover the constraint velocity $\dot{\vct x}_c$ and the free space velocity $\dot{\vct x}_f$ from \eqref{eq:qd2}, the inverse of $\vct S$ such that $\vct S ^{-1} \vct S=\vct I_{n\times n}$ is easily found utilizing the above properties and \eqref{eq:AZ} to be:
\begin{equation}
    \vct S^{-1} = \begin{bmatrix}
    \vct A\\
 \vct Z^{{\dagger}^\mathrm{T}} \end{bmatrix}\in \mathbb{R}^{n\times n}.
\end{equation} 
Thus, the following forward kinematic mapping holds: 
\begin{equation}\label{eq:xc_xf}
    \begin{bmatrix}
    \dot{\vct x}_c \\
    \dot{\vct x}_f
    \end{bmatrix}
    =\begin{bmatrix}
    \vct A\\
 \vct Z^{{\dagger}^\mathrm{T}} \end{bmatrix}\dot{\vct q}_d.
\end{equation}
which  decouples RCM constrained and unconstrained motion since it satisfies
% \eqref{eq:xc_xf} are   decoupled if 
the following properties for all nonzero $\vct{\dot x}_f$ and $\vct{\dot x}_c$ \cite{park1999dynamical}: % which is easily established  using   properties \eqref{eq:AZ} and  (ii):
\begin{equation*}
\vct A \vct S
\begin{bmatrix}
    \vct 0\\
     \dot{\vct x}_f
    \end{bmatrix} =\vct 0, \  \vct Z^{{\dagger}^\mathrm{T}} \vct S
\begin{bmatrix}
    \dot{\vct x}_c \\
    \vct 0
    \end{bmatrix} =\vct 0.
\end{equation*}
Hence, the velocity in the free space does not affect the velocity in the RCM constraint space and vice versa.
\added{\begin{remark}Notice that the proposed mapping between the joint velocities $\dot{\vct q}_d$ and the free motion coordinates $\dot{\vct x}_f$ in eq \eqref{eq:xc_xf} differs from that of \cite{kastritsi21TRMB},  which utilizes the $\vct Z \vct W $ matrix instead of the $\vct Z^{{\dagger}^T}$ transpose of this work.  The proposed mapping enables passivity to be proved in the presence of repulsive potentials which is lost if the  mapping of \cite{kastritsi21TRMB} is utilized.\end{remark}}

\subsection{Target admittance model}
We are now ready to propose the following  target dynamics to achieve all our control objectives:
\begin{equation} \label{eq:qddot_2}\begin{split}
 \ddot{{\vct{q}}}_d+\vct S \begin{bmatrix}
     \vct  h\\
     \vct  u 
    \end{bmatrix} = \vct Z^\mathrm{T} \vct Z \vct J_t^\mathrm{T} \bigg(\vct F_{th}+\vct{F}_r\bigg) 
 \end{split} 
\end{equation} 
 where \begin{equation} \label{eq:HU} \begin{split}
    \vct  h=\dot{\vct A}\dot{\vct q}_d+2\alpha\dot{\vct x}_c+\beta^2 \vct x_c, \ \  \vct  u=\bigg(\vct D_f \vct Z^{{\dagger}^\mathrm{T}}   +\dfrac{d(\vct Z^{{\dagger}^\mathrm{T}} )}{dt}\bigg)\dot{\vct q}_d
\end{split} 
\end{equation}
with $ \vct D_f \in \mathbb{R}^{(n-2)\times (n-2)} $ being  a diagonal matrix of positive damping gains, $  \alpha, \ \beta$ are  positive 
gains, $\vct F_{r}=\begin{bmatrix}
    {\vct f}_r\\
     \vct 0_{3\times 1} 
    \end{bmatrix}\in \mathbb{R}^{6 \times 1}$ is the total repulsive force at the tip, and $\vct F_{th} \in \mathbb{R}^{6}$ is the human generalized 
force transformed  in the tool-tip. In particular, if  the user exerts the generalized force $\vct F_h \in \mathbb{R}^{6}$, at the end-effector (tool basis) 
then 
$ \vct F_{th}=\vct T_{te}  \vct F_h$
with $ \vct T_{te} =\begin{bmatrix}
    \ \ \  \vct I_{3\times 3} \ \ \ \ \ \ \vct 0_{3\times 3} \\
    (\vct p_e-\vct p_t)^\wedge   \ \  \vct I_{3\times 3} 
    \end{bmatrix} 
$ where $\vct p_e$ is the position of the end-effector w.r.t. the base frame. The selection of gains is detailed in the rest of this section.
Target dynamics \eqref{eq:qddot_2}, \eqref{eq:HU} are integrated to produce joint motion references $\vct q_d, \dot{\vct q}_d$ to the robot which is assumed to faithfully reproduce them. The next theorem and its proof describes the main result of the proposed admittance controller.

\begin{theorem}  \label{theorem1}
 The following statements for the target dynamics \eqref{eq:qddot_2}, \eqref{eq:HU}  with initial joint position $\vct q_d(t_0) $ such that $\vct p_t(t_0) \notin\mathcal{O}_c $ are valid:
 \begin{enumerate}
     \item  the dynamics of the RCM-constraint space are  decoupled from the dynamics in the free space.
     \item $\dot{\vct{x}}_c$, ${\vct{x}}_c$ converge  exponentially  to zero.
     \item the system  is strictly output passive with respect to the  velocity $\dot{\vct{x}}_f$,  under the exertion of a generalized human force 
     \item the forbidden region is never violated
 \end{enumerate}

\begin{proof}

Taking the time derivative of $\dot{\vct x}_c$ from \eqref{eq:xc_xf}
and substituting $\ddot{\vct q}_d$ from \eqref{eq:qddot_2}, yields the constraint motion dynamics:
\begin{equation} \label{eq:xcddot}
\begin{split}
 \ddot{\vct x}_c=-\vct A \vct S \begin{bmatrix}
     \vct  h\\
     \vct  u 
    \end{bmatrix} + \vct A\vct Z^\mathrm{T} \vct Z \vct J_t^\mathrm{T} \bigg(\vct F_{th}+{\vct{F}_r} \bigg)  +\dot{\vct A}\dot{\vct q}_d .
  \end{split}
\end{equation}
Utilizing property (i) and  \eqref{eq:AZ} it is easy to show that  $\vct A \vct S=[\vct I_{2\times 2}  \  \ \vct 0_{2\times (n-2)} ] $. Substituting $ \vct  h$  from \eqref{eq:HU} in \eqref{eq:xcddot} yields the exponentially stable constraint motion dynamics:
\begin{equation} \label{eq:xc_ddot_}
 \ddot{\vct x}_c+2\alpha \dot{\vct x}_c+\beta^2 {\vct x}_c=0.
\end{equation}
which ensure that ${\vct x}_c$ and $\dot{\vct x}_c$ will converge exponentially to zero completing the proof of theorem's statement (2). It is now clear that  the $\vct h$  term in \eqref{eq:qddot_2} is instrumental in yielding the above linear exponentially stable system for the constraint space motion. Parameters $\alpha$ and $\beta$ are equal for a critically damped response and their value determines the speed of exponential convergence of $\vct x_c$ to zero.
Notice that the set:
\begin{equation}
{ {\mathcal {S}}\triangleq \left\lbrace (\vct  q_d , \dot{\vct q}_d)\in \mathbb {R} ^{2n}\mid \dot{\vct x}_c=0, \vct x_c=0 \right\rbrace } \end{equation} is  positively invariant. In other words, once a trajectory of the system starts or enters ${\mathcal {S}}$ it will evolve within this set for all times.  Within this set,  \eqref{eq:qd2} becomes  $\dot{\vct q}_d=\vct Z^\mathrm{T} \dot{\vct x}_f$; left multiplied by $\vct J_t$ yields the generalized tip velocity 
${\vct v}_t=\vct Z_x^\mathrm{T} \dot{\vct x}_{f_{\{1:4\}}}$ which clarifies the physical meaning of the free motion velocity coordinates $\dot {\vct x}_{f}$; the first is   
a linear velocity in the direction of the tool axis, the next three correspond to the angular velocity while the remaining $n-6$ coordinates refer to the robot's redundant dof.

 Taking the time derivative of  $\dot {\vct x}_f$ from \eqref{eq:xc_xf} and substituting $\ddot{\vct q}_d$ from \eqref{eq:qddot_2}, yields:
 
\begin{equation*}\begin{split} \ddot{{\vct{x}}}_f =-\vct Z^{{\dagger}^\mathrm{T}}\vct S \begin{bmatrix}
     \vct  h\\
     \vct  u 
    \end{bmatrix} +\vct Z^{{\dagger}^\mathrm{T}}\vct Z^\mathrm{T} \vct Z \vct J_t^\mathrm{T} \bigg(\vct F_{th}+{\vct{F}_r} \bigg) +\dfrac{d(\vct Z^{{\dagger}^\mathrm{T}} )}{dt}\dot{\vct q}_d.\end{split}
\end{equation*}

Utilizing properties (ii) and (iii) it is easy to show that $\vct Z^{{\dagger}^\mathrm{T}} \vct S=[\vct 0_{(n-2)\times 2}  \  \ \vct I_{(n-2)\times (n-2)} ] $. Thus, substituting   $ \vct  u$ from \eqref{eq:HU}   yields  the free space motion dynamics:
\begin{equation} \label{eq:free_motion_dyn}
\begin{split}
 \ddot{\vct x}_f + \vct D_f \dot{\vct x}_f=\vct Z \vct J_t^\mathrm{T} \bigg(\vct F_{th}+{\vct{F}_r} \bigg)  
\end{split}
\end{equation}

\added{
\begin{remark} Notice that  $\vct Z\vct J_t^\mathrm{T} = \begin{bmatrix}
     \vct Z_x 
   \\
    \vct 0_{(n-6)\times 6}
    \end{bmatrix}$. Hence any forces multiplied by $\vct Z_x$ \eqref{eq:Zx} results in forces along the tool axis and torques around the entry port. Thus in \eqref{eq:free_motion_dyn} there is a correspondence between the free motion velocity coordinates  and  the filtered force components $\vct Z_x (\vct F_{th}+{\vct{F}_r} ) $ that leads to an intuitive motion under the exerted force. In fact, the first component of $\dot {\vct x}_f$ is a linear velocity in the direction of the tool axis and the next three components correspond to the angular velocity. This natural correspondence is lacking from \cite{kastritsi21TRMB}. 
\end{remark}
}

It is clear that $\vct D_f$ introduces damping along the coordinates of $\vct {\dot{x}}_f$. Damping gains for the angular velocity coordinates $\dot{\vct x}_{f_{\{2:4\}}}$
should be equal for a synchronized response.
  Thus proof of statement (1) regarding the decoupling of  free motion dynamics \eqref{eq:free_motion_dyn}  from the constraint dynamics \eqref{eq:xc_ddot_} is completed.

For the proof of statement (3) consider
the following candidate Lyapunov function:
\begin{equation}\label{V_q2}
V_f=\dfrac{1}{2}\dot {\vct x}_f^\mathrm{T}\dot {\vct x}_f+ \sum_{\vct p_i \in \mathcal{O}_s} {V}_i(\vct p_t).
\end{equation}
Its time derivative along  the set $\mathcal {S}$ by substituting $\ddot{\vct x}_f$ from \eqref{eq:free_motion_dyn}, is given by:
 \begin{equation}\label{eq:vfdot}
 \begin{split}
     \dot V_f=&-\dot {\vct x}_f^\mathrm{T} \vct D_f  \dot {\vct x}_f+\dot {\vct x}_f^\mathrm{T} \vct Z \vct J_t^\mathrm{T} \bigg(\vct F_{th}+{\vct{F}_r} \bigg)  -{\vct{F}_r}^\mathrm{T}  {\vct v}_t.
 \end{split}
\end{equation} 
Substituting $ {\vct v}_t$  from \eqref{eq:Jq} yields:
\begin{equation}
 \begin{split}
     \dot V_f&=-\dot {\vct x}_f^\mathrm{T} \vct D_f  \dot {\vct x}_f+\dot {\vct x}_f^\mathrm{T}\vct Z \vct J_t^\mathrm{T} \bigg(\vct F_{th}+{\vct{F}_r} \bigg)  - {\vct{F}_r}^\mathrm{T}  \vct J_t\dot {\vct q}_d.
 \end{split}
\end{equation} 
Furthermore, substituting $\dot {\vct q}_d$  from \eqref{eq:qd2} in the invariant set $\mathcal {S}$ yields:
% \begin{equation}\label{eq:vf2dot}
%  \begin{split}
%      \dot V_f=-\dot {\vct x}_f^\mathrm{T} \vct D_f  \dot {\vct x}_f+\dot {\vct x}_f^\mathrm{T} \vct Z \vct J_t^\mathrm{T} \bigg(\vct F_{th}+\begin{bmatrix}\vct f_r
%     \\ \vct 0_{3\times 1}
%  \end{bmatrix} \bigg) \\- \begin{bmatrix}\vct f_r
%     \\ \vct 0_{3\times 1}  \end{bmatrix}^\mathrm{T}  \vct J_t \vct Z^\mathrm{T} \vct{\dot x}_f .
%  \end{split}
% \end{equation} 
% Thus, \eqref{eq:vf2dot} becomes:
\begin{equation}\label{eq:vf3dot}
 \begin{split}
     \dot V_f=-\dot {\vct x}_f^\mathrm{T} \vct D_f  \dot {\vct x}_f+\dot {\vct x}_f^\mathrm{T}   \vct Z\vct J_t^\mathrm{T}\ \vct F_{th} 
 \end{split}
\end{equation} 
which implies that the output  $\dot {\vct x}_f$  is strictly passive  under the exertion  of  \added{the filtered} human  generalized force \added{$\vct Z\vct J_t^\mathrm{T}\ \vct F_{th}$.}

   Rewriting \eqref{eq:vf3dot} by completing the squares, yields:
 	\begin{equation}\label{vdot_last_squares}
 	\begin{split}
 	\dot{V}_f=&-||\sqrt{\vct{D}_f}\dot{\vct{x}}_f-\dfrac{1}{2}\sqrt{\vct{D}_f}^{-1}\vct Z\vct J_t^\mathrm{T}\ \vct F_{th} ||^2\\&+\dfrac{1}{4}  \vct F_{th}^\mathrm{T}\vct J_t \vct Z^\mathrm{T}  \vct D_f^{-1} \vct Z\vct J_t^\mathrm{T}\ \vct F_{th} \leq \dfrac{1}{4}  \vct F_{th}^\mathrm{T}\vct J_t \vct Z^\mathrm{T}  \vct D_f^{-1} \vct Z\vct J_t^\mathrm{T}\ \vct F_{th} 
 	\end{split}
 	\end{equation} 
	Notice that $\vct{F}_h$ represents the force applied by the human to guide the robot. Thus, $\vct{F}_h$ is bounded and therefore $\vct{F}_{th} $ and  $\vct F_{th}^\mathrm{T}\vct J_t \vct Z^\mathrm{T}  \vct D_f^{-1} \vct Z\vct J_t^\mathrm{T}\ \vct F_{th}$ are also bounded functions of time. Additionally, the human forces have bounded energy. Hence integrating   \eqref{vdot_last_squares} we get:
\begin{equation}\label{int}
V_f\leq V_f(t_0) + \int \dfrac{1}{4}  \vct F_{th}^\mathrm{T}\vct J_t \vct Z^\mathrm{T}  \vct D_f^{-1} \vct Z\vct J_t^\mathrm{T}\ \vct F_{th} < \infty.
\end{equation}
Thus, $ {V}_i(\vct p_t)$ and $ \dot{{\vct{x}}}_f$ are bounded under the exertion of human force.   As a consequence, there exists a positive constant $\overline{\varepsilon}_i$ such that: $ V_i \leq \overline{\varepsilon}_i$. The boundedness of  $ V_i$ given by \eqref{vi} implies that the tip will never violate the forbidden area since
  $\|\vct{p}_t - \vct{p}_i\|\geq d_1>d_c$  completing the proof of statement (4). 
 
 \end{proof}\end{theorem}

The above theoretical justification is extended to the case the forbidden area should not be violated by the whole tool in the following subsection.

% \begin{remark}

\begin{figure}[!h]
  \centering
  \includegraphics[width=0.9\linewidth]{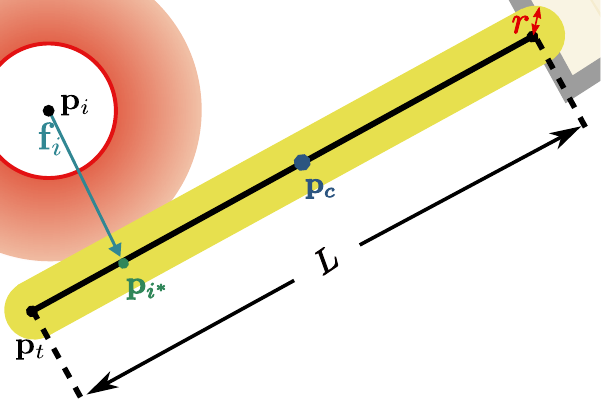}
  \caption{ The nearest point on the capsule axis $\vct p_i^*$.}
  \label{fig:whole}
\end{figure}

% \end{remark}}
{\subsection{Extension to the Whole Tool Body}\label{extention}}
% In many cases, RCM satisfaction is desirable along with the restriction of part of the  body or  
To extend the analysis from the tool tip to the whole tool or a tool segment we  model  the  tool  body  as  a  capsule with $r \in \mathbb{R}^+$ being its  radius, thus  allowing  the analytical computation of the point on the capsule axis $\vct p_i^* \in \mathbb{R}^3$  with minimum distance from a point $\vct p_i \in \mathcal{O}_s$ in the forbidden region cloud
(see Figure \ref{fig:whole}).
We then impose a repulsive force on it.

Given the length $L$ of the capsule axis which is a line segment, every point belonging to it is given by:
\begin{equation}
\vct{p}_s (\sigma ;t)=\vct p_t(t)-\vct n_t  L \sigma
\label{eq:ps}
\end{equation}
 with $\sigma \in [0 , 1]$.
Then at every control cycle and for each point $\vct p_i \in \mathcal{O}_s$, we find the nearest point  $\vct{p}{^*_i} =\vct{p}_s ( {\sigma}{^*_i},t)$  on the line segment  of the capsule  where ${\sigma}{^*_i}$ can be found analytically as described  in our previous paper \cite{kastritsi2019manipulation} :
\begin{equation} \label{eq:s_star}
  \sigma{^*_i} = 
      \begin{cases*}
       \zeta_i, & \text{if  $0 \leq \zeta_i \leq 1$}\\
        1, & \text{if $\zeta_i > 1$} \\
        0, & \text{if  $\zeta_i < 0$}
        \end{cases*}
\end{equation}
where 
\begin{equation*}
  \zeta_i=\dfrac{1}{L}\vct n_t^\mathrm{T}(\vct p_t -\vct p_i).
\end{equation*}
 
Then the barrier artificial potential   $V_i(\vct p^*_i)$  \eqref{vi}  can be applied for each  pair $(\vct p_i, \vct p^*_i)$. Notice that in this case  $d_c$ in \eqref{vi} is replaced by $d_c+r$.

 To impose actively the constraint for the capsule a repulsive force is applied at $\vct{p}^*_i$ calculated by the negative gradient of  $ V_i({\vct p^*_i}) $, i.e.,  $ \vct{f}_i(\vct p{^*_i})=-\dfrac{\partial V_i}{\partial \vct p{^*_i}}$  \eqref{eq:f} for all points of the cloud.
This is transformed in the tool-tip as a  force $\vct{f}_i$ and a torque $\bm \tau_i$ given by: 
 \begin{equation}\label{eq:fiti}
     \begin{bmatrix}
    {\vct f}_{i}\\
     \bm \tau_{i} 
    \end{bmatrix}= \begin{bmatrix}
    \ \ \ \ \vct I_{3\times 3} \ \ \ \ \ \vct 0_{3\times 3} \\
    (\vct p_i^*-\vct p_t)^\wedge \ \ \vct I_{3\times 3}
    \end{bmatrix}\begin{bmatrix}
   {\vct f}_i \\
  \vct 0_{3\times 1}
    \end{bmatrix}.\in \mathbb{R}^{6 \times 1},
   \end{equation}The total generalized repulsive force on the tool-tip is given by the sum of the $\vct{f}_i$ forces and torques  $\bm{\tau}_i$ produced by each artificial potential field i.e., \begin{equation}\label{eq:sum2}
    \vct{F}_r =  \sum_{\vct p_i \in \mathcal{O}_s} \begin{bmatrix}
   {\vct f}_i \\
   {\bm \tau}_i
    \end{bmatrix}.
\end{equation}
Then \eqref{eq:sum2} is provided to the target admittance model \eqref{eq:qddot_2}.
The proof of stability and enforcement of RCM and forbidden area constraints in this case is similar to the case of the tool-tip. 

The proof of the first two statements of Theorem \ref{theorem1} are equivalent. To prove the  third and fourth statement for the case of whole tool body constraint  the following candidate Lyapunov is utilized:
\begin{equation}
V_f=\dfrac{1}{2}\dot {\vct x}_f^\mathrm{T}\dot {\vct x}_f+\sum_{\vct p_i \in \mathcal{O}_s} {V}_i({\vct p^*_i}).    
\end{equation}  Taking its time derivative yields: 
\begin{equation}\label{eq:vdotwhole}
\dot{V}_f=-\dot {\vct x}_f^\mathrm{T} \vct D_f  \dot {\vct x}_f+\dot {\vct x}_f^\mathrm{T} \vct Z \vct J_t^\mathrm{T} \bigg(\vct F_{th}+{\vct{F}_r} \bigg)-\sum_{\vct p_i \in \mathcal{O}_s}\vct{f}_i(\vct p{^*_i})^\mathrm{T}\dot{\vct p}_i^* .\end{equation}
Furthermore, utilizing the time derivative of the nearest point 
$\dot{\vct p}_i^*=  \begin{bmatrix}
   \ \vct I_{3\times 3} \ \ \  (\vct p_t-\vct p_i^*)^\wedge  
    \end{bmatrix}\mathbf v_t- \vct n_t  L \dot{{\sigma}{^*_i}}$ in \eqref{eq:vdotwhole} yields: 
    \begin{equation}\label{eq:vdot_whole}
 \begin{split}
     \dot V_f=&-\dot {\vct x}_f^\mathrm{T} \vct D_f  \dot {\vct x}_f+\dot {\vct x}_f^\mathrm{T} \vct Z \vct J_t^\mathrm{T} \bigg(\vct F_{th}+{\vct{F}_r} \bigg)  +\sum_{\vct p_i \in \mathcal{O}_s}\vct{f}_i(\vct p{^*_i})^\mathrm{T}( \vct n_t  L \dot{{\sigma}{^*_i}})\\&-\sum_{\vct p_i \in \mathcal{O}_s}\vct{f}_i(\vct p{^*_i})^\mathrm{T}\begin{bmatrix}
   \ \vct I_{3\times 3} \ \ \  (\vct p_t-\vct p_i^*)^\wedge  
    \end{bmatrix}\mathbf v_t.
 \end{split}
\end{equation}
    Utilizing $\vct{F}_r$ from \eqref{eq:sum2} and using \eqref{eq:fiti} yields:
    \begin{equation}\label{eq:vdot_whole2}
 \begin{split}
     \dot V_f=&-\dot {\vct x}_f^\mathrm{T} \vct D_f  \dot {\vct x}_f+\dot {\vct x}_f^\mathrm{T} \vct Z \vct J_t^\mathrm{T} \bigg(\vct F_{th}+{\vct{F}_r} \bigg)  -{\vct{F}_r}^\mathrm{T}  {\vct v}_t\\&+\sum_{\vct p_i \in \mathcal{O}_s}\vct{f}_i(\vct p{^*_i})^\mathrm{T}( \vct n_t  L \dot{{\sigma}{^*_i}}).
 \end{split}
\end{equation}
 For the first case of \eqref{eq:s_star} utilizing \eqref{eq:ps} for $\sigma={\sigma}{^*_i}$
 yields that $\vct n_t^\mathrm{T}(\vct p{^*_i}-\vct p_i)=0$ 
 and for the other two cases $\dot{\sigma}^*_i=0$. As $\vct{f}_i(\vct p{^*_i})$ is in the direction of $(\vct p{^*_i}-\vct p_i)$ we get:
 \begin{equation}
 \begin{split}
     \dot V_f=&-\dot {\vct x}_f^\mathrm{T} \vct D_f  \dot {\vct x}_f+\dot {\vct x}_f^\mathrm{T} \vct Z \vct J_t^\mathrm{T} \bigg(\vct F_{th}+{\vct{F}_r} \bigg)  -{\vct{F}_r}^\mathrm{T}  {\vct v}_t.
 \end{split}
\end{equation}
The above equation is equivalent to \eqref{eq:vfdot} in the proof for the  tool-tip case, thus the proof of the two last statements follows the same line.

\begin{figure}[!h]
  \centering
  \subfloat[The velocity-dependent term $ \mathrm D_{v_i}$]{\includegraphics[width=0.43\linewidth]{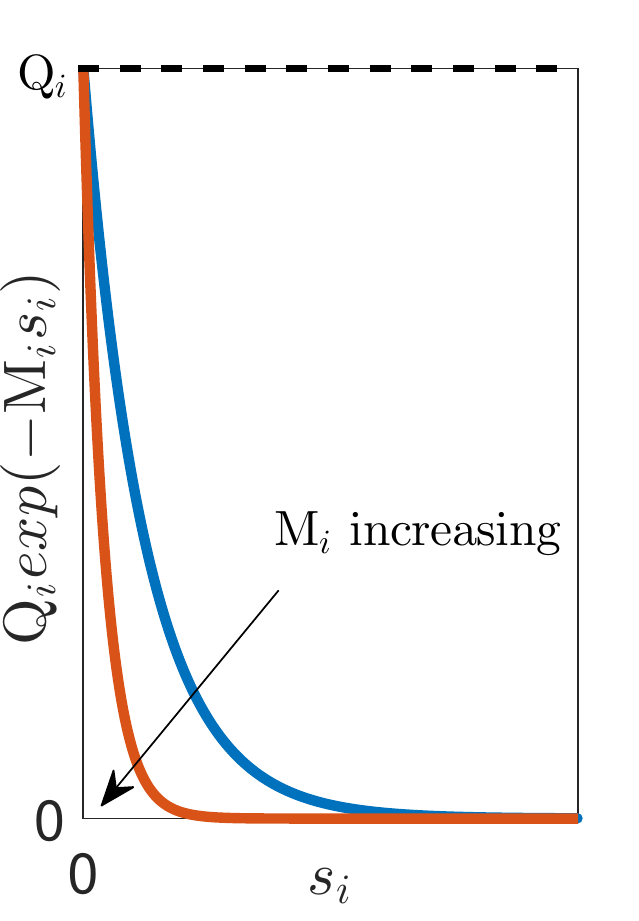}}\hspace*{0.2cm} 
  \subfloat[The term  $ \mathrm D_{r_i}$ depending on the repulsive force]{\includegraphics[width=0.43\linewidth]{ 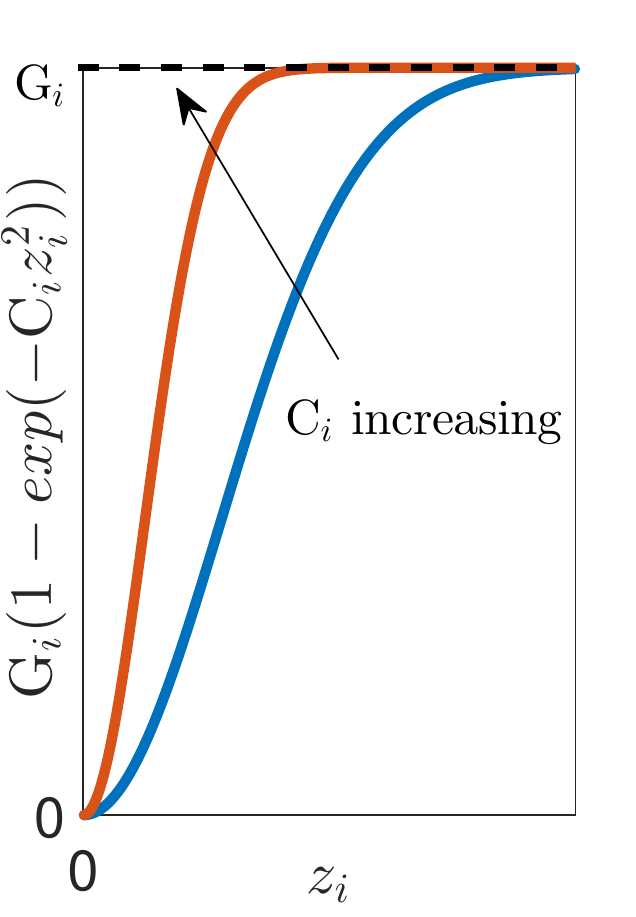}%
  }
  \caption{The variable damping terms. }
  \label{fig:damping}
\end{figure}

\added{\subsection{Variable Damping }\label{damping}}
In order to facilitate kinesthetic guidance variable damping gains have been proposed in the literature \cite{ficuciello2015variable}. In our case, the need for variable damping is more acute, as near the forbidden areas the manipulation performance of the tool is  affected by the non-linear increase of the apparent  stiffness induced by the artificial potentials. In order to achieve the objective of smooth performance in the whole of the tool's workspace we utilize variable damping gains. In particular,   the following variable damping gains are utilized in $\vct D_f \in \mathbb{R}^{(n-2)\times (n-2)} $:
\begin{equation}
    \mathrm D_{f_i}=\mathrm D_{c_i}+\mathrm D_{v_i}+\mathrm D_{r_i}, \  \ i=1, ...,n-2
\end{equation} 
where $\mathrm D_{c_i}$ are positive constant values given for  $i=1, ..., (n-2)$  and  $\mathrm D_{v_i}$ and $\mathrm D_{r_i}$ are variable damping terms added in the first four diagonal elements to facilitate motion along the task unconstrained coordinates. The term $\mathrm D_{v_i}$ is  inspired by \cite{ficuciello2015variable} and is given by : 
\begin{equation}
    \mathrm D_{v_i}=\begin{cases*}
        \mathrm Q_i exp(-\mathrm  M_i s_i), & \text{for  $i=1, ...,4$}\\
        0, & \text{for  $i=5, ...,n-2$}
        \end{cases*}
\end{equation}  where $\mathrm Q_{i}$ and $\mathrm M_i$ are positive constant gains, $s_1=|\dot{ x}_{f_{1}}|$ and $s_j=||\dot{\vct x}_{f\{2:4\}}||$ for $ j=2,3,4$. This exponential term provides high damping at low velocities and low damping for high velocities (see Fig \ref{fig:damping} a) in order to facilitate the user to easily displace the tool in free space. 
The term $\mathrm D_{r_i}$ serves to enhance the protection of the forbidden region and the feeling of control over the task  within the range of influence of the repulsive field.
It provides increased damping as the virtual stiffness induced by the artificial potential increases. Thus unwanted oscillatory motion is avoided. The term $\mathrm D_{r_i}$ is given by:
 \begin{equation}
  \mathrm   D_{r_i}=
        \begin{cases*}
       \mathrm G_{i}(1-exp(-\mathrm C_i z_i^2)), & \text{for  $i=1, ...,4$}\\
        0, & \text{for  $i=5, ...,n-2$}
        \end{cases*}
 \end{equation}  where $\mathrm C_i, \ \mathrm G_{i}$ are constant gains, $z_1=|F_{rx\{1\}}|$ and  $z_j=||{\vct F}_{rx\{2:4\}}||$ for $ j=2,3,4$ {where $\vct F_{rx}=\vct Z_x^\mathrm{T} \vct F_r$}. This term utilizes the magnitude of the repulsive force along the tool axis and the magnitude of repulsive torques. It increases the damping when these forces are high and is zero beyond the influence region.  (see Fig \ref{fig:damping} b).  Notice that  $\mathrm D_{v_i}$ and $\mathrm D_{r_i}$ are bounded non-negative functions thus they do not affect the stability proof.
 
\textbf{\\}

\section{Experimental results}\label{section:experiment}
\begin{figure}[!h] \label{}
  \centering 
  \subfloat[]{\includegraphics[width=0.20\textwidth]{ 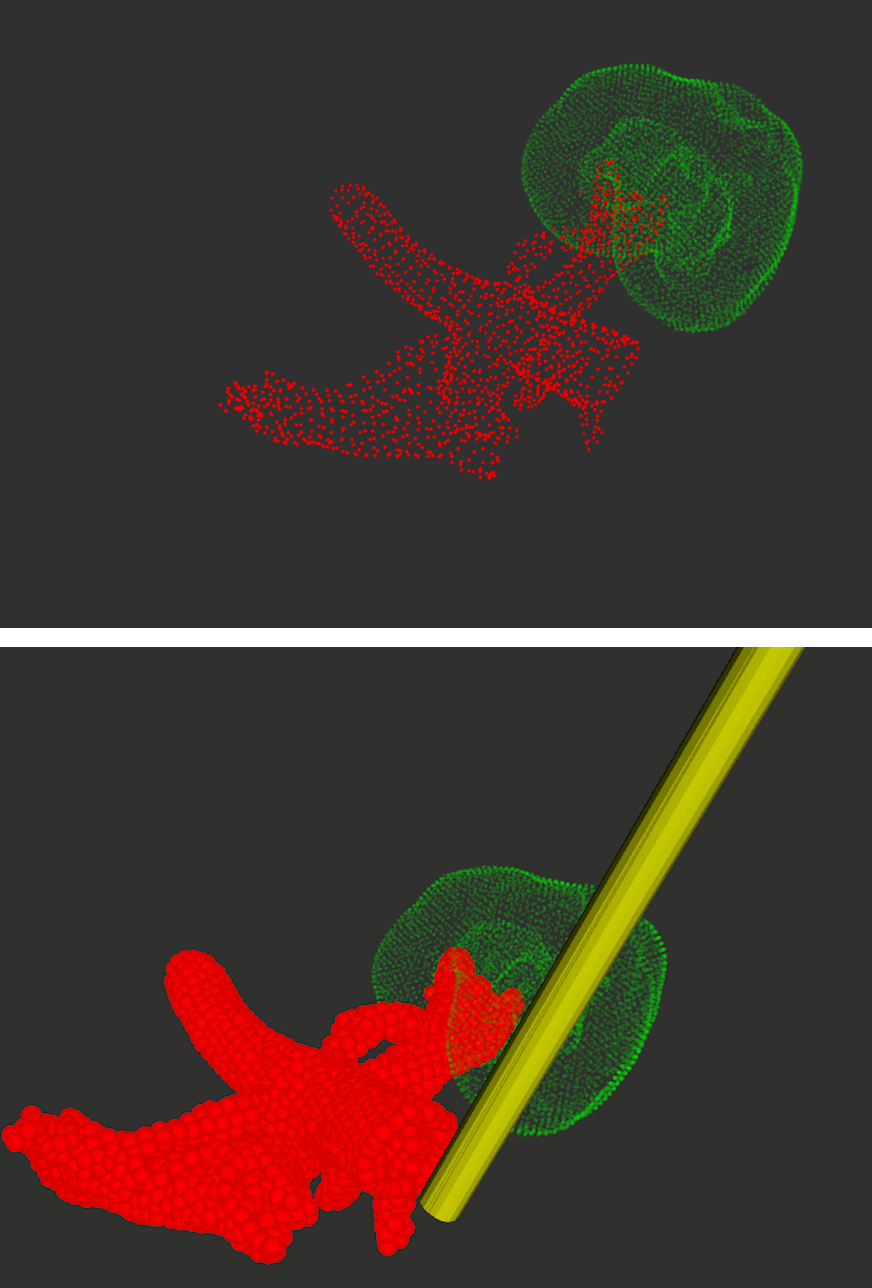}%
  }\hspace*{0.01cm} 
  \subfloat[]{\includegraphics[width=0.25\textwidth]{ 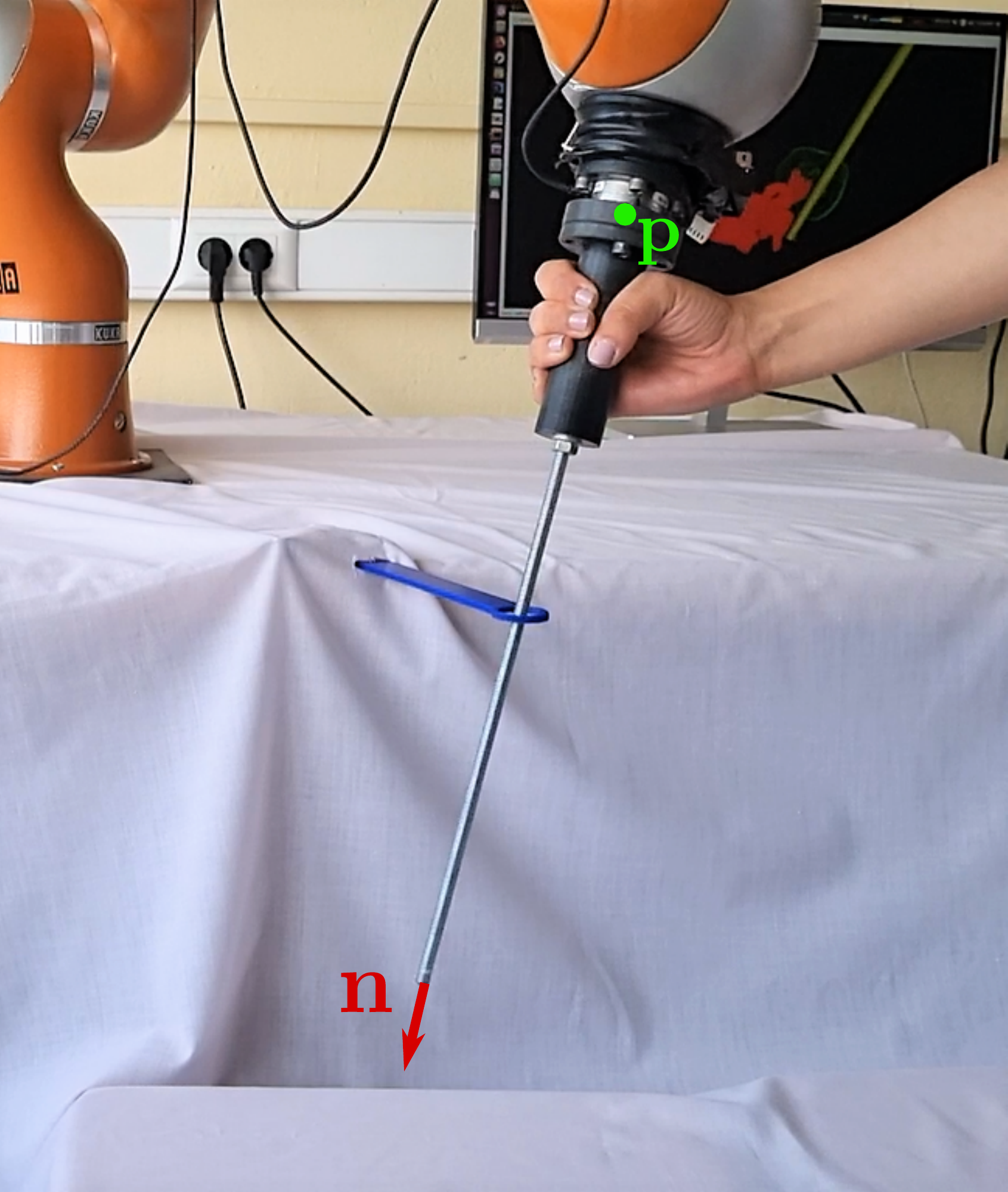}}
  \caption{ \added{Experimental setup: (a) Visualization  of  the  point  cloud  of  a  kidney  (green)  and  its surrounding  vessels  (red)  as the forbidden one with the overlapping spheres of the forbidden region shown in the bottom. (b) The robot holds a long tool mimicking a surgical instrument in a real  RAMIS hands-on procedure.}}
  \label{fig:exp_setup}
\end{figure}

In order to validate the effectiveness and performance of the proposed  target admittance model,  a 7-dof KUKA LWR4+ robotic manipulator with a force/torque sensor (ATI Mini40) attached at its end-effector is used in  joint position control mode. This mode of operation provides a high bandwidth control of the robot dynamics so that the desired joint motion $\vct q_d$ from \eqref{eq:qddot_2} is faithfully tracked. Hence,  approximately $\vct q=\vct q_d$.   The set-up is mimicking a hands-on surgical procedure. In particular, the robot holds a tool of length  $0.43 \ m$ and diameter  $7 \ mm$, mimicking  a  surgical instrument in  a  real  RAMIS procedure. Typically the diameter of a surgical instrument is in the range of  3-12 mm. The entry port, imitating the incision point on the patients body wall, is a ring softly attached to the environment with the position of its center being $\vct p_c=[-0.6053 \ \  -0.2203  \ \ 0]^\mathrm{T} \ m$. The initial joint configuration is  $\vct q_d(0)=[ 20 \   50   \ 0     \   -70  \  0   \  60  \ 0]^\mathrm{T}\  deg$ so that the tool passes through  the central point of the ring  (Figure  \ref{fig:exp_setup}).  In a real surgical procedure its position can be found utilizing the robot \cite{Ortmaier2000}, \cite{Dong16}, \cite{Gruijthuijsen18} or camera images from the external surgical scene \cite{Rosa2015}.

% The ring is time invariant as We exclude motion from the breathing function. 
 \begin{table}[!t]
	\caption{Parameters of the proposed method. }
	\centering
	\begin{tabular}{|p{1.58cm}|p{1.58cm}|p{1.58cm}|p{1.58cm}| }
 \hline
 	$\vct W $ &	$\alpha, \	\beta $   &$\mathrm d_0 $  & 	$\mathrm k_i $ \\
 \hline
    $1.5 \mathrm I_{7\times 7}$ &$25$ &  $ 0.0115 \ m $ &    $ 0.01$ \\
 \hline	
\end{tabular}
	 
	\begin{tabular}{ |p{1.4cm}||p{0.8cm}|p{0.8cm}|p{0.8cm}|p{0.8cm}|p{0.8cm}| }
 \hline
 \multicolumn{6}{|c|}{Parameters for the variable Damping matrix  $\mathbf D_f$} \\
 \hline
 Element $ i$ &$\mathrm D_{c_i}$  &$\mathrm Q_i$ & $\mathrm M_i$&$\mathrm G_i$ & $\mathrm  C_i$\\
 \hline
 $1$   & 10    & 25 &  22&   60&   0.01\\
$\{2:4\}$ &   4  & 20   & 19 &   30&   0.2\\
 $5$ &60 & -&  -&   -&   -\\
 \hline	
\end{tabular}
	
	\label{FIG:TAB}
\end{table} 
The point cloud of an internal human organ (kidney with green points) and its surrounding vessels (red points) is imported in the robot's workspace (Figure  \ref{fig:exp_setup}). The task is to manipulate the tool in the surgical area of the kidney avoiding the sensitive region of the surrounding vessels.   After down sampling the  point  cloud  of  the  vessels  to  increase computational performance, the radius of the spheres to cover  the  empty  spaces  is  calculated  to  be $ \mathrm d_c=3.5mm$. 
In  the  real  case  the  3D  point  cloud  of  the surgical site will be provided by the endoscopic camera and the characterization of the forbidden areas can be performed by the surgeon via a friendly human machine interface. For providing visual feedback to the user, a virtual scene, displaying the point cloud of the surgical site and the virtual timestamp of the real robot with the tool, is created utilizing Rviz embedded in ROS framework (Figure  \ref{fig:exp_setup}).  The whole scheme is implemented in C++ with control  frequency $\mathrm{f_s} = 250$ using the \added{ Fast Research Interface (FRI) library. The FRI library is provided by the manufacturer and allows the communication between the KUKA LWR4+ Robot's Controller (KRC) and a remote PC over the Ethernet \cite{schreiber2010fast}.} \added{The target admittance model \eqref{eq:qddot_2} with the parameters values given in table \ref{FIG:TAB} is integrated every 4ms using the Euler method  and the derivative of the matrix $\mathbf{A}$ used in the model is numerically calculated using the backward differentiation formula. } 

\added{For the experimental evaluation,} the user \added{was asked to guide} %\deleted{guides}
the tool  on the surgical area and towards the sensitive region entering the region of  active constraints influence. Experiments demonstrate RCM manipulation and that the forbidden region is never violated by the tool tip (first experiment) and  by the whole tool (scecond experiment).
\subsection{ {Tool}-tip case}

\begin{figure}[!htbp]
  \centering
    \subfloat[\added{Traces of the {tool} axis} ]{\includegraphics[width=0.45\textwidth]{ 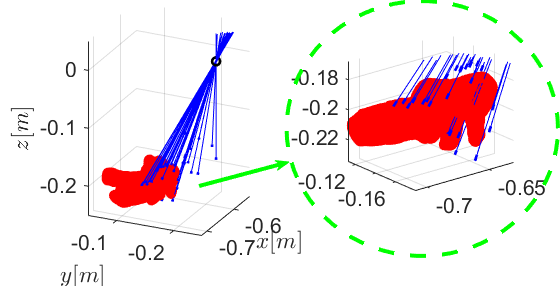}}
    \par
  \subfloat[The minimum distance between $\vct p_c$ and the tool axis $\vct n$ where  $\vct p^*=\vct p+\vct  n \vct n^\mathrm{T}(\vct p_c-\vct p)$.]{\includegraphics[width=0.37\textwidth]{ 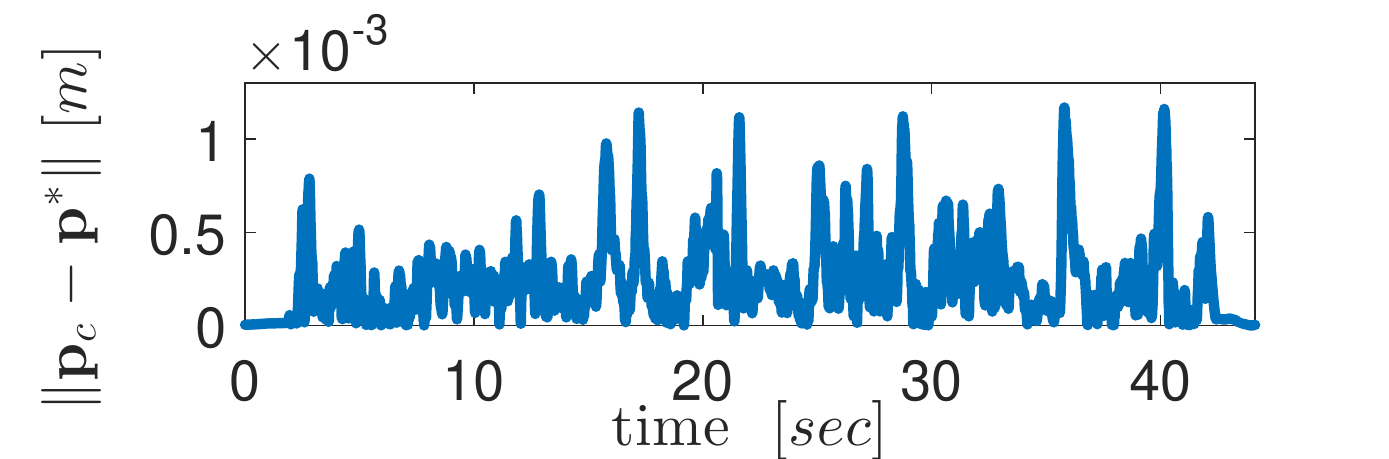}%
  }\par
  \subfloat[The minimum distance between the tip of the tool and the point cloud of the sensitive area.]{\includegraphics[width=0.37\textwidth]{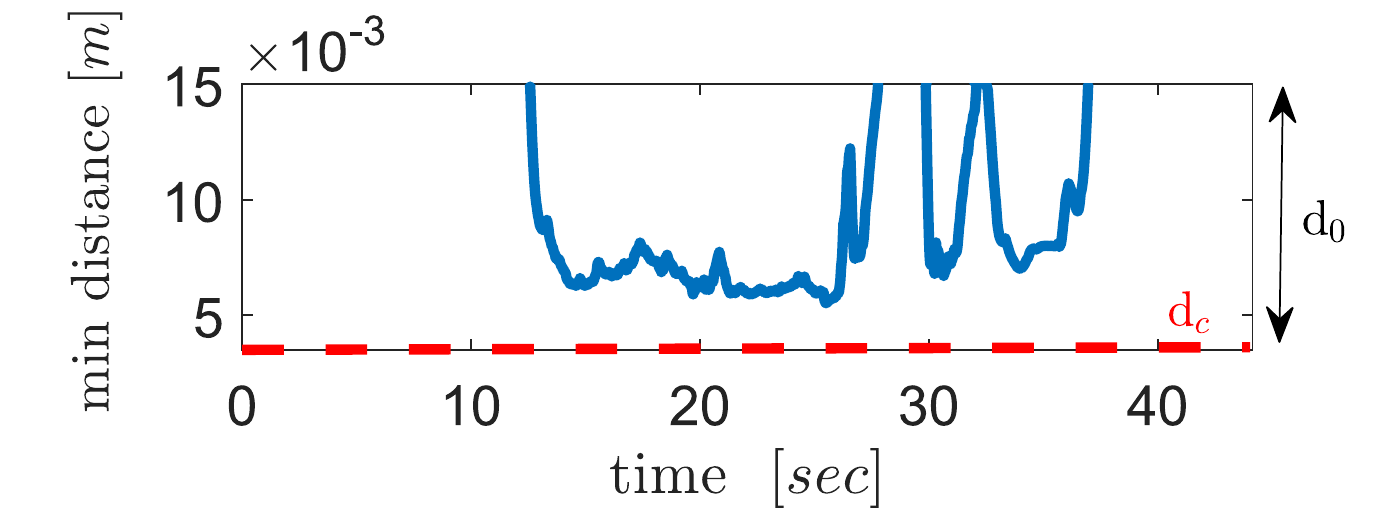}%
  }\par
  \subfloat[Repulsive forces and torques.]{\includegraphics[width=0.37\textwidth]{ 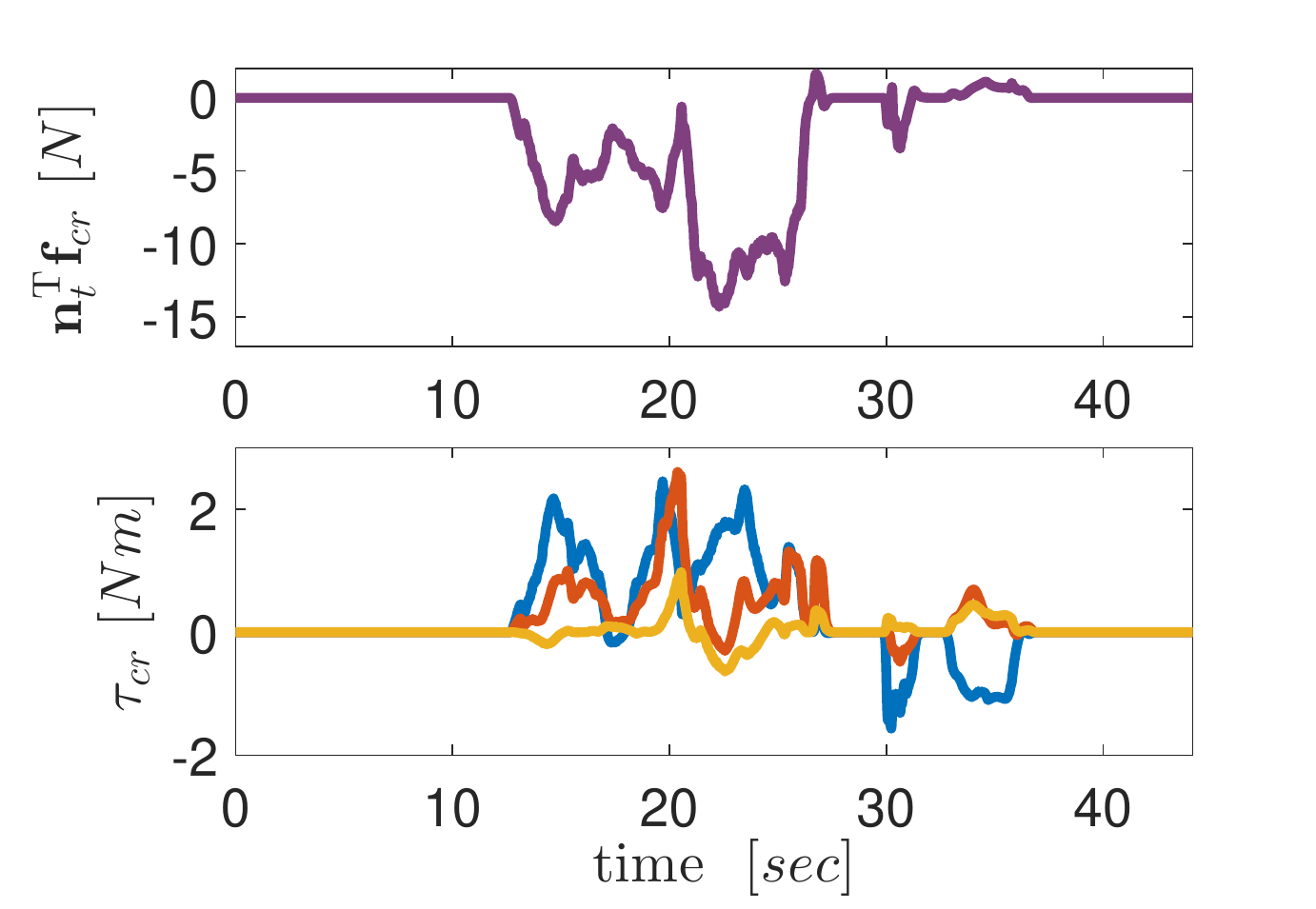}}
  \par
  \subfloat[Human force and torque.]{\includegraphics[width=0.37\textwidth]{ 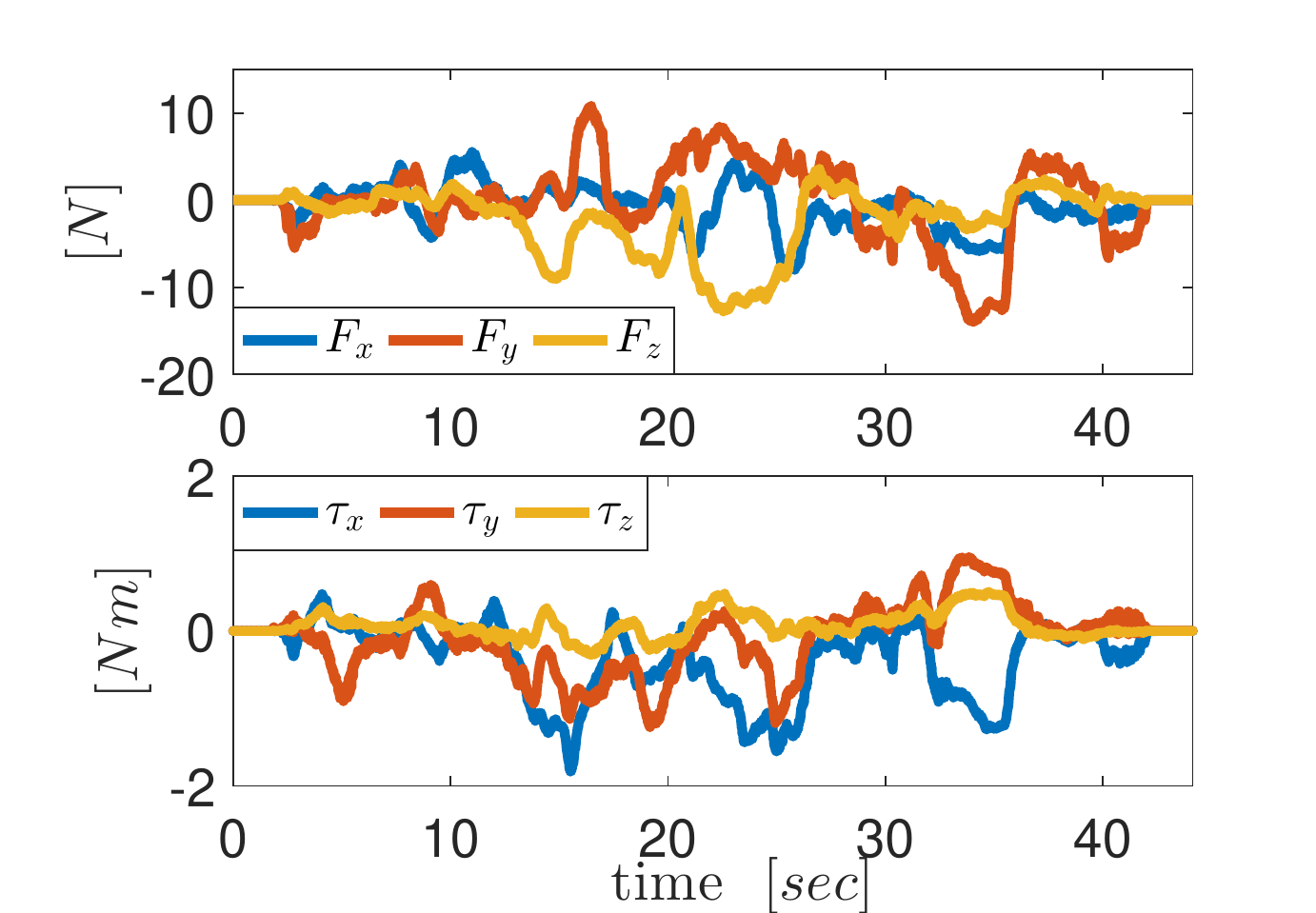}%
  }
  \caption{Experimental results: case of {tool}-tip constraint.}
  \label{fig:exp_results}
\end{figure}

\begin{figure}[!htbp]
  \centering
  \subfloat{\includegraphics[width=0.38\textwidth]{ 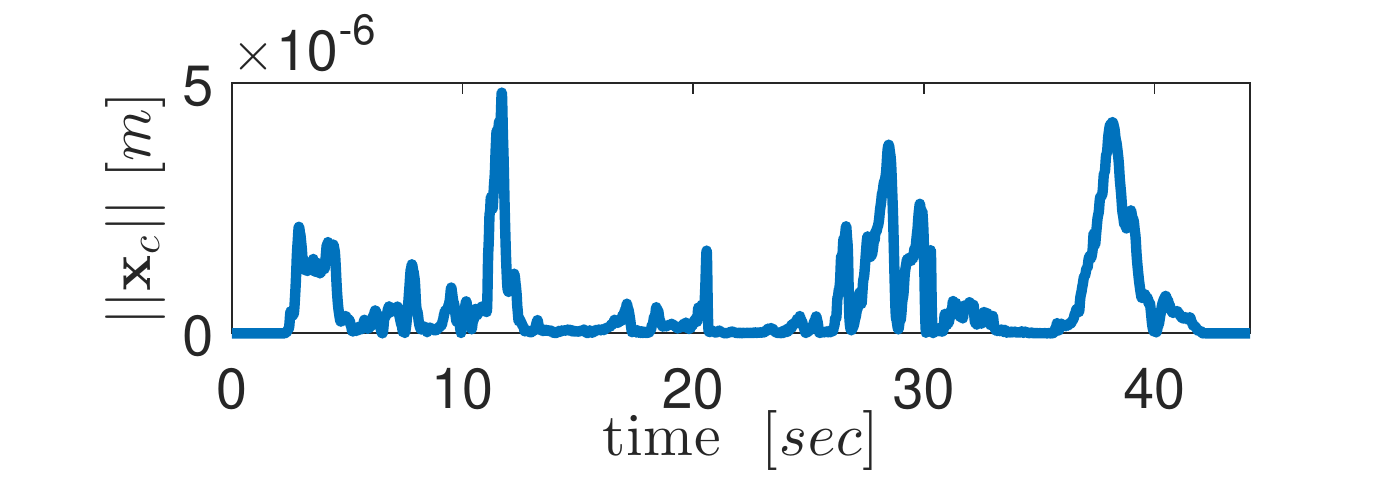}}
  \caption{The norm of the desired  position  constraint $||\vct x_c||$. }
  \label{fig:xc}
\end{figure}

\begin{figure}[!htbp]
  \centering
  \subfloat{\includegraphics[width=0.38\textwidth]{ 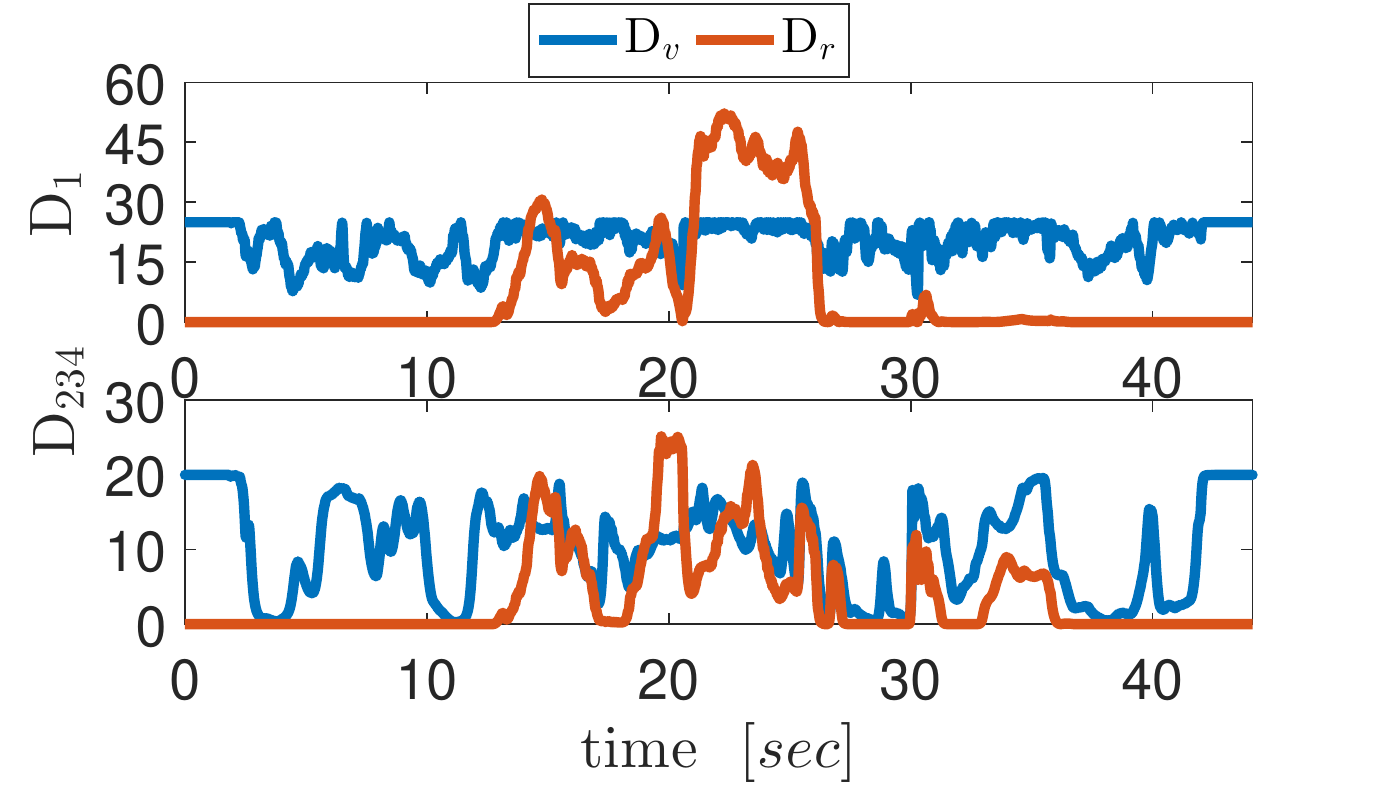}}
  \caption{Damping gain variation during the experiment.}
  \label{fig:dam_tip}
\end{figure}

The user guides the tool-tip initially on the surgical area and then approaches the sensitive region (vessels)  entering the region of the active constraints influence. Experimental results are presented in Figure  \ref{fig:exp_results}. 
 Human  exerted  forces and torques are shown Figure  \ref{fig:exp_results}-e.  After $t_1$=12.5 sec the user  has guided the {tool}-tip at a distance less than $d_0+d_c=0.015m$ from the vessels triggering the generation of repulsive forces at the tip. Figure  \ref{fig:exp_results}-d shows repulsive forces and torques acting on the entry port i.e. $\vct n_t ^\mathrm{T} {\vct f}_{cr}$ and ${\bm\tau }_{cr}$ respectively where $ \begin{bmatrix}
    {\vct f}_{cr}\\
     \bm \tau_{cr}
    \end{bmatrix}= \begin{bmatrix}
    \ \ \ \ \vct I_{3\times 3} \ \ \ \ \ \vct 0_{3\times 3} \\
    (\vct p_t-\vct p_c)^\wedge \ \ \vct I_{3\times 3}
    \end{bmatrix} \vct F_r \in \mathbb{R}^{6 \times 1}$. These forces actively resist the user for damaging the sensitive region with the tool tip.
   This is confirmed by the tip distance from the forbidden area shown in Figure  \ref{fig:exp_results}-c after $t_1$=19 sec which remains greater than $d_c$. \added{Figure  \ref{fig:exp_results}-a shows the traces of the tool during the manipulation}.
  Clearly, the desired RCM constraint is satisfied as also shown in Figure  \ref{fig:exp_results}-b  by the minimum  distance  between  the incision point  $\vct p_c$ and the axis of the {tool},  $\vct n$ 
  %where $\vct p^*=\vct p+\vct  n \vct n^\mathrm{T}(\vct p_c-\vct p)$, 
  which is kept less than $1.2 mm$;  
    ideally this distance should stay within the range of displacement $\vct x_c$ in the constraint space whose norm is shown in Figure \ref{fig:xc} and is in the order of $10^{-6} m$. The higher displacement $1.2mm$ in the experiment can 
be attributed  to the fact that the controlled robot is not infinitely stiff hence   robot joint positions are approximately equal but not identical to the provided reference values. Clearly the controlled robot stiffness affects the magnitude of errors we get regarding the satisfaction of the desired RCM constraint. \added{In Figure \ref{fig:exp_results}-a  the embedded subplot shows details of the tool traces where it can be observed that part of the tool are in some instances inside the forbidden area. This is expected as in this experiment the avoidance concerns only the tool tip.} 
    {Figure  \ref{fig:dam_tip} shows the varying damping gains during the experiment.  $\mathrm{D}_{v_i}$  are shown in blue and  $\mathrm{D}_{r_i}$ in red. Notice in the beginning and in the end of the motion the terms $\mathrm{D}_{v_i}$ that have  maximum value as there is zero velocity. Further notice how $\mathrm{D}_{r_i}$ take non-zero values when entering the region of influence of the repulsive potential and how they  increase when respective repulsive force/torques increase. }

\subsection{ Whole {tool} case}
\begin{figure}[!htbp]

  \centering
    \subfloat[\added{Traces of the {tool} axis.} ]{\includegraphics[width=0.45\textwidth]{ 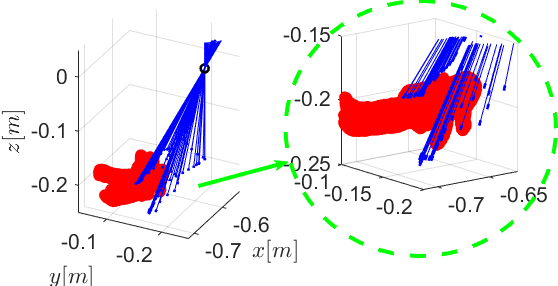}}
    \par
  \subfloat[The minimum distance between $\vct p_c$ and the tool axis $\vct n$  where  $\vct p^*=\vct p+\vct  n \vct n^\mathrm{T}(\vct p_c-\vct p)$.]{\includegraphics[width=0.37\textwidth]{ 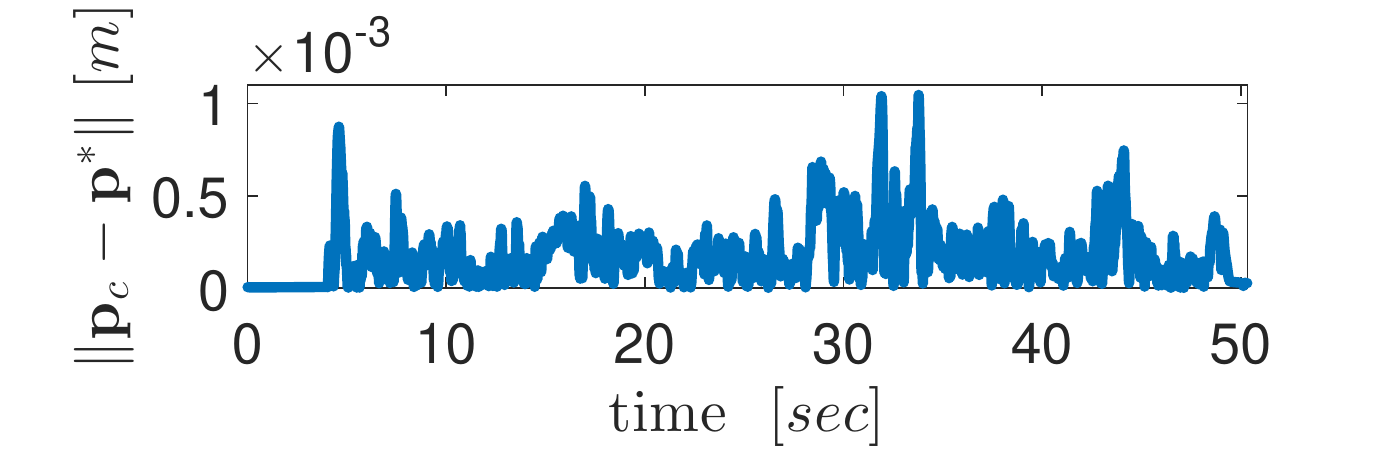}%
  }\par
  \subfloat[The minimum distance between the tool axis  and the point cloud of the sensitive area.]{\includegraphics[width=0.37\textwidth]{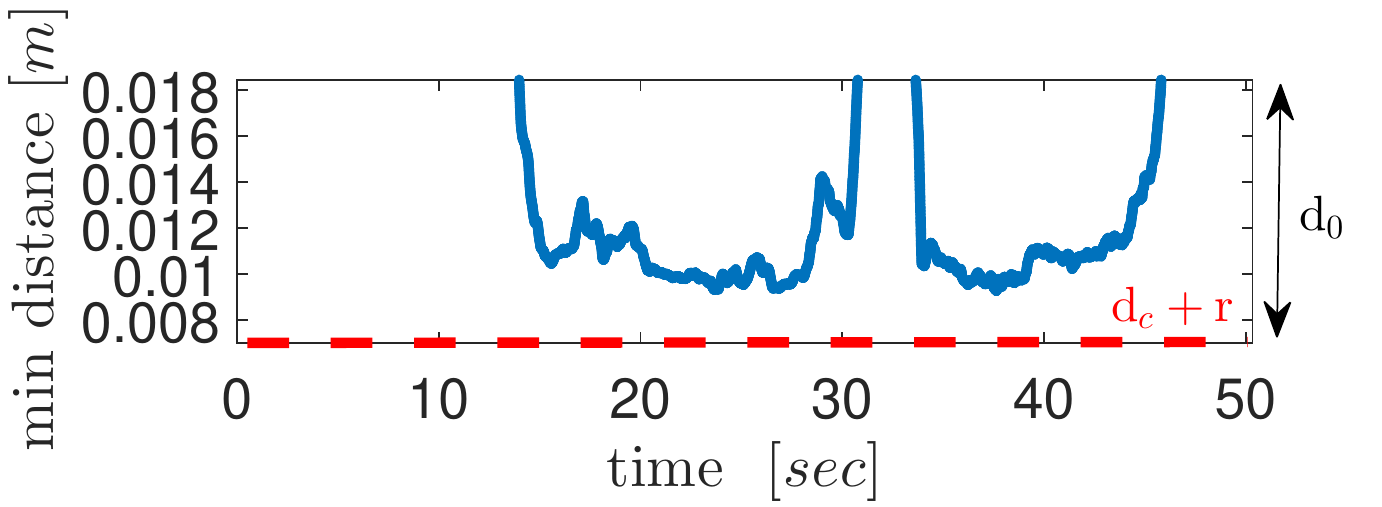}%
  }\par
  \subfloat[Repulsive forces and torques.]{\includegraphics[width=0.37\textwidth]{ 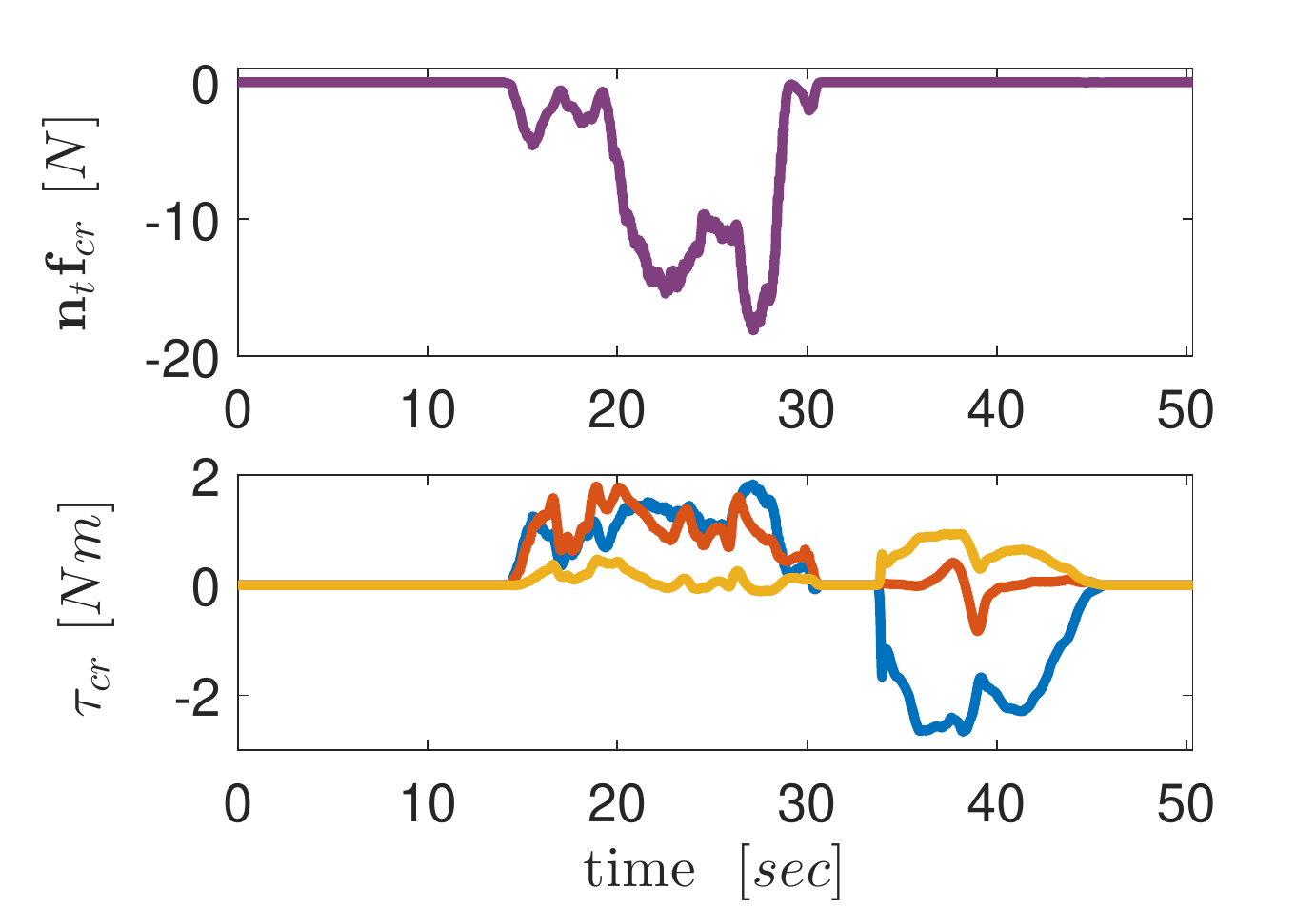}}
  \par
  \subfloat[Human force and torque.]{\includegraphics[width=0.37\textwidth]{ 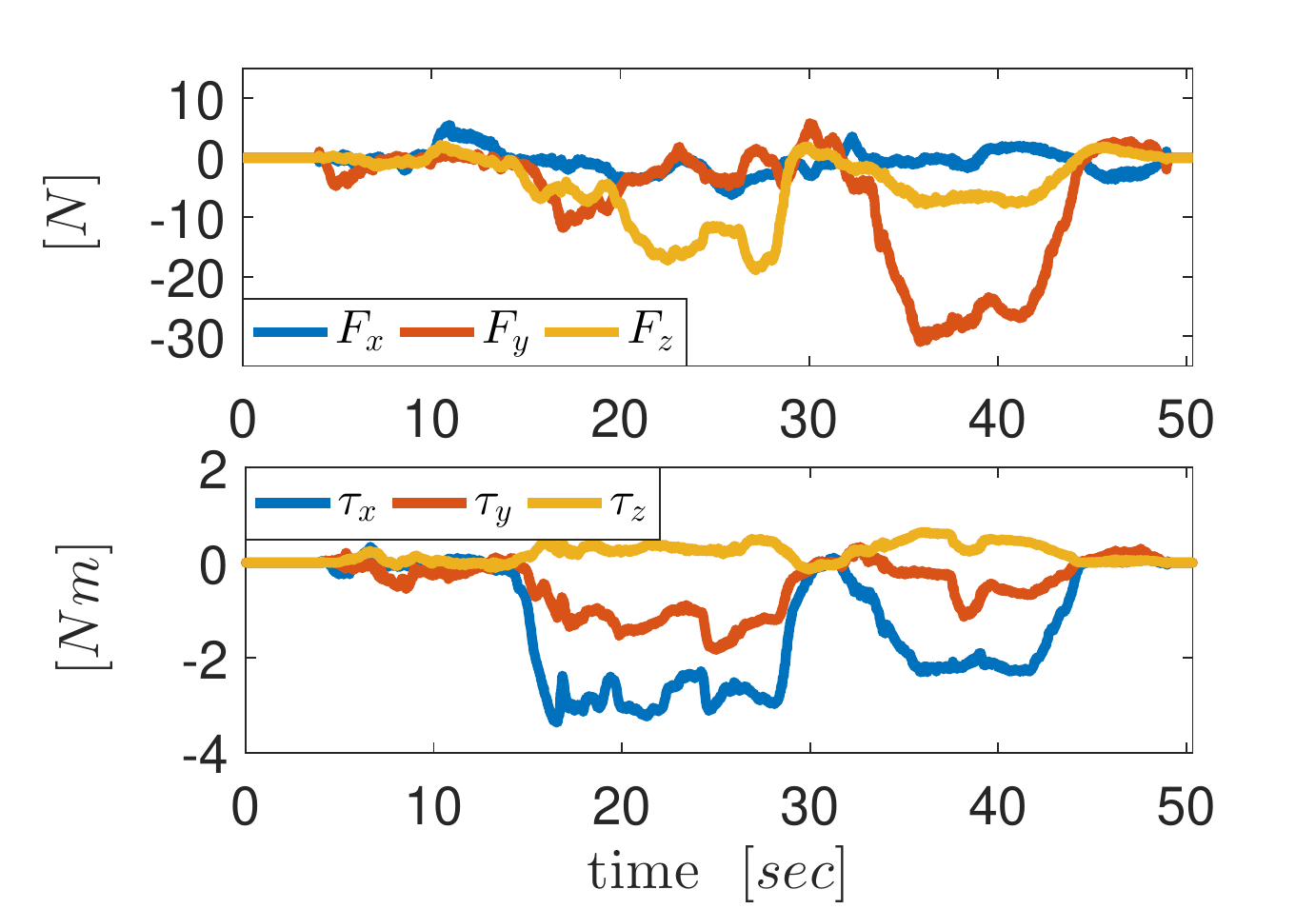}%
  }
  \caption{Experimental results: case of whole {tool} constraint. }
  \label{fig:exp_results_2}
\end{figure}

\begin{figure}[!htbp]
  \centering
  \subfloat{\includegraphics[width=0.38\textwidth]{ 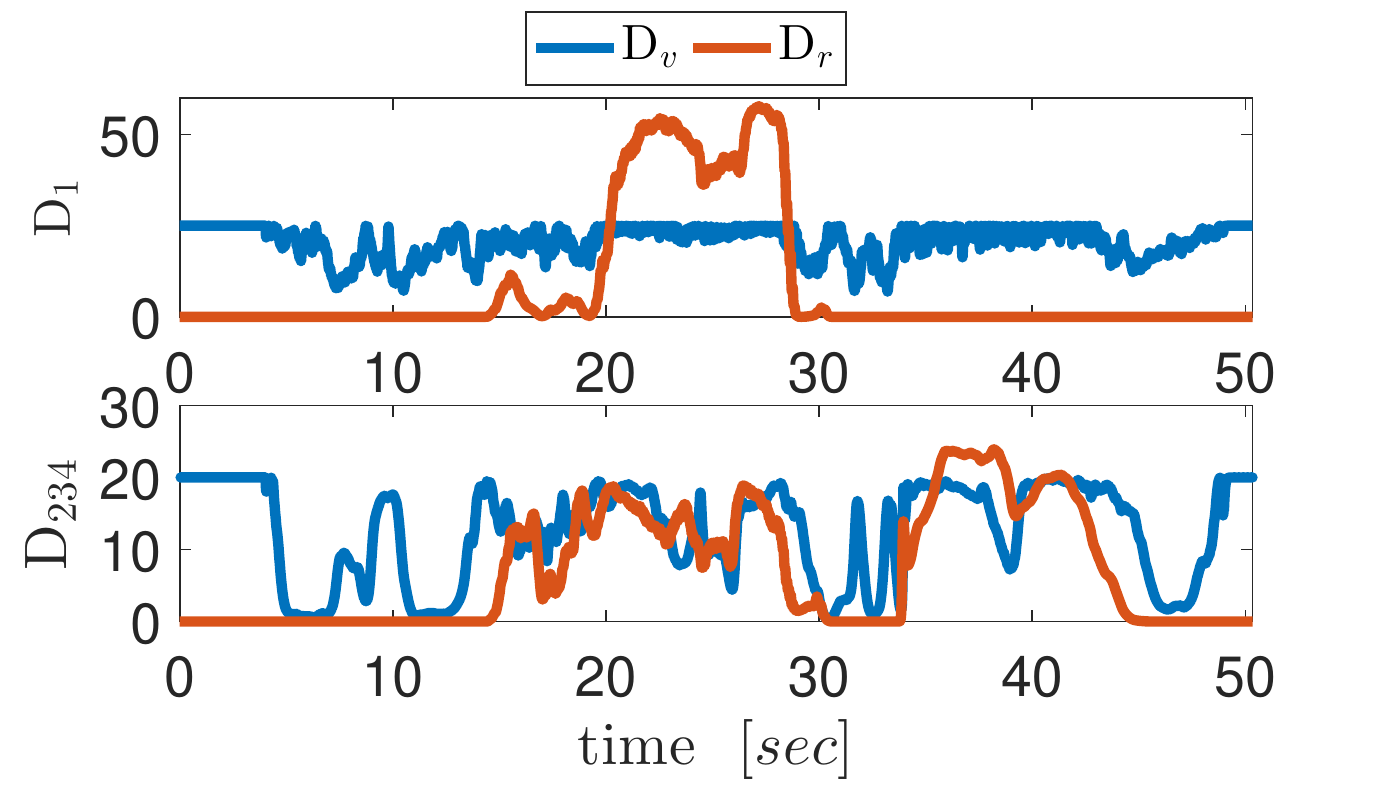}}
  \caption{Damping gain variation during the experiment. }
  \label{fig:dam_tool}
\end{figure}

A capsule with radius $\mathrm r=3.5 mm$ encloses the elongated tool. As in the previous case the user  guides the {tool}  towards to  the sensitive region (vessels)  entering the region of the active constraints influence. Experimental results are presented in Figure  \ref{fig:exp_results_2}.
 Human  exerted  forces and torques are shown in Figure  \ref{fig:exp_results_2}-e. \added{Notice that the user applied  forces of magnitude close to 30 N, with the system responding in a stable smooth manner. Such large forces were exerted on purpose close to the forbidden areas to investigate the controller's performance but much smaller forces are adequate during surgery.}  After $t_1=13$ sec the user  has guided the {tool} at a distance less than $d_0+d_c+r=0.0185m$ from the vessels triggering the generation of repulsive forces and torques shown in Figure  \ref{fig:exp_results_2}-d that actively resist the user for damaging the sensitive region as shown by the minimum {tool} distance from the forbidden area (Figure  \ref{fig:exp_results_2}-c after $t_1=13$ sec) which remains greater than $d_c+r$. \added{Figure \ref{fig:exp_results_2}-a shows the traces of the tool during the manipulation. Clearly, the desired RCM constraint is satisfied as
also shown in Figure \ref{fig:exp_results_2}-b by the minimum distance between the
incision point $\vct p_c$ and the axis of the tool, $\vct n$ which is kept less
than 1.1mm as shown in 
Figure  \ref{fig:exp_results_2}-b.
In Figure \ref{fig:exp_results_2}-a  the embedded subplot shows details of the tool traces where it can be observed that the whole tool never violates the forbidden area as expected .} 
{Figure  \ref{fig:dam_tool} shows the varying damping gains.} The  video  of  the  experiment  can  be  seen  in:   \href{https://youtu.be/NB1x8WzwbOQ}{ $ https://youtu.be/NB1x8WzwbOQ $}. 

The proposed admittance control scheme have been validated via experiments involving tools with different lengths. We expected that the parameter values of $ G_i $ and $ C_i, i = \{2: 4 \} $ involved in the   variable damping term $ \mathbf D_f $ may need to be adjusted because the torques induced on the RCM by the repulsive forces are amplified in the case of longer instruments. However, keeping the same control parameter values we found that the performance was satisfactory
when the tool length was in the range of 30 cm to 43 cm.

\section{Conclusions}\label{section:conclution}
This paper proposes a passive  admittance controller achieving RCM and spatial constraint satisfaction to guarantee the safety of sensitive regions. The admittance model is designed in such a way so that the control action that  guarantees manipulation away from the spatial constraints does not affect the satisfaction of the RCM constraint and vice versa. \added{Moreover, it
provides manipulation transparency  and smooth motion even close to the forbidden regions enhancing the user's sense of control over the task}. Experimental results on 7-dof KUKA LWR4+  robot manipulator simulating a real minimally invasive surgery procedure demonstrate the efficacy of the proposed scheme. The  whole tool never touches the forbidden region and tool manipulation via a RCM is achieved in a intuitive manner. Future work focus on adapting the proposed method for a tele-operated setup \added{and testing it in a real surgical environment}.
  \bibliographystyle{model1-num-names}
\bibliography{references.bib}

\end{document}